\documentclass{article} 
\usepackage[nonatbib,preprint]{neurips_2024}

\usepackage{amsmath,amsfonts,bm}

\def\eqref#1{equation~\ref{#1}}

\def\1{\bm{1}}

\DeclareMathAlphabet{\mathsfit}{\encodingdefault}{\sfdefault}{m}{sl}
\SetMathAlphabet{\mathsfit}{bold}{\encodingdefault}{\sfdefault}{bx}{n}

\newcommand{\R}{\mathbb{R}}

\usepackage[utf8]{inputenc} 
\usepackage[T1]{fontenc}    
\usepackage{hyperref}       
\usepackage{url}            
\usepackage{booktabs}       
\usepackage{amsfonts}       
\usepackage{nicefrac}       
\usepackage{microtype}      
\usepackage{xcolor}         

\usepackage{hyperref}
\usepackage{url}

\usepackage{algorithm}
\usepackage{algorithmicx}
\usepackage[noend]{algpseudocode}
\usepackage{amsmath, amsfonts, amssymb, amsthm, amsbsy, amscd, bm, bbm,mathrsfs}   
\usepackage{graphicx}
\usepackage{subcaption}
\usepackage{cleveref}
\usepackage{enumitem}
\usepackage{titlesec}
\usepackage{wrapfig}
\usepackage{multirow}
\usepackage{booktabs}

\usepackage{graphicx}
\graphicspath{{figures/}}
\numberwithin{equation}{section}

\theoremstyle{definition}

\theoremstyle{definition}

\renewcommand{\d}{\bm{d}}

\newcommand{\g}{\mathbf{g}}
\newcommand{\y}{\bm{y}}

\renewcommand{\b}{\mathbf{b}}

\newcommand{\I}{\mathbf{I}}

\newcommand{\G}{\mathbf{G}}

\renewcommand{\R}{\mathbb{R}}
\newcommand{\w}{\bm{w}}

\newcommand{\A}{\mathbf{A}}

\renewcommand{\P}{\mathbf{P}}

\renewcommand{\H}{\mathbf{H}}

\title{A Hessian-informed hyperparameter optimization for differential learning rate}

\author{Shiyun Xu\\
University of Pennsylvania\\
\texttt{shiyunxu@sas.upenn.edu} \\
\And
Zhiqi Bu\thanks{This work does not relate to ZB’s position at Amazon.} \\
Amazon\\
\And
Yiliang Zhang\\
DRW
\And
Ian Barnett \\
University of Pennsylvania
}

\begin{document}

\maketitle

\begin{abstract}
    Differential learning rate (DLR), a technique that applies different learning rates to different model parameters, has been widely used in deep learning and achieved empirical success via its various forms. For example, parameter-efficient fine-tuning (PEFT) applies zero learning rates to most parameters so as to significantly save the computational cost. 

    At the core, DLR leverages the observation that different parameters can have different loss curvature, which is hard to characterize in general. We propose the Hessian-informed differential learning rate (Hi-DLR), an efficient approach that solves the hyperparameter optimization (HPO) of learning rates and captures the loss curvature for any model and optimizer adaptively. Given a proper grouping of parameters, we empirically demonstrate that Hi-DLR can improve the convergence by dynamically determining the learning rates during the training. 

\end{abstract}

\section{Introduction}
\textit{Differential learning rate} (DLR) is a technique that assigns different learning rates to distinct parameter groups. Here the parameter groups are partitions of model parameters $\w\in\R^D$ such that $\w=[\w_{(1)},...,\w_{(K)}]$ and the gradient $\g=[\g_{(1)},...,\g_{(K)}]$. 

When $K=1$ for a single parameter group, this reduces to the update of uniform learning rate (ULR) and we update with a scalar $\eta_t$ such that
$$\w_{t+1}=\w_t-\eta_t\g_{t}.$$
When $K>1$ for multiple parameter groups, the model is updated by multiple learning rates $\eta_{(k)}$ such that
$$\w_{t+1}=\w_t-[\eta_{(1)}\g_{(1),t},...,\eta_{(K)}\g_{(K),t}]:=\w_t-\bm\eta_{[K],t}\g_{[K],t}$$

In deep learning, DLR can plays an important role for complex model architectures and tasks, as it allows for more precise control over the pace at which different parts of a model learn.

\begin{figure}[!htb]
    \centering
\includegraphics[width=0.33\linewidth]{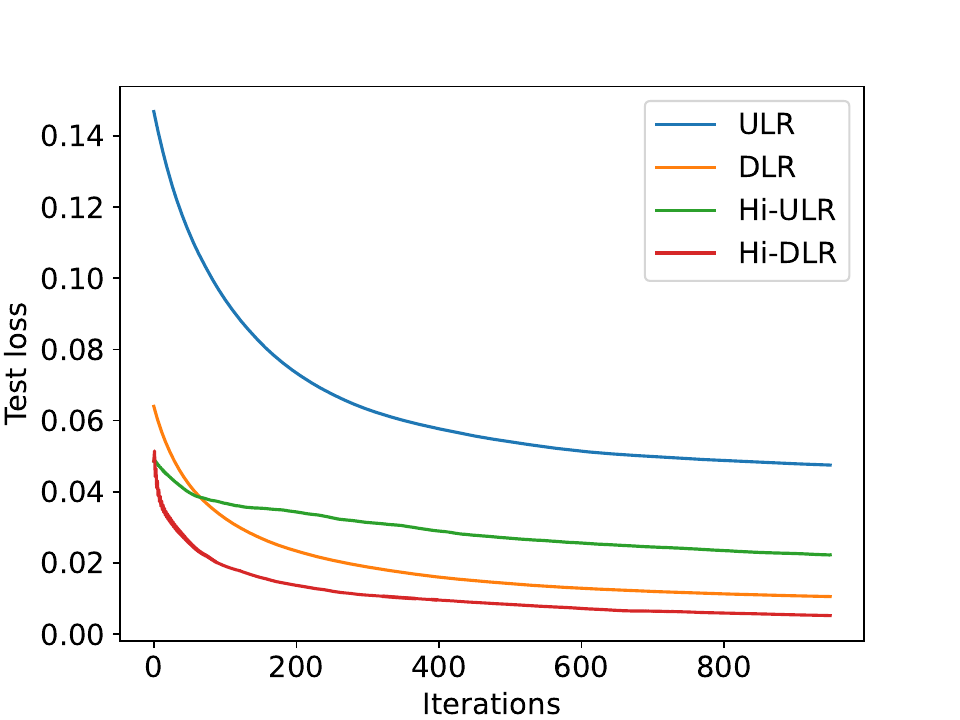}
\includegraphics[width=0.31\linewidth]{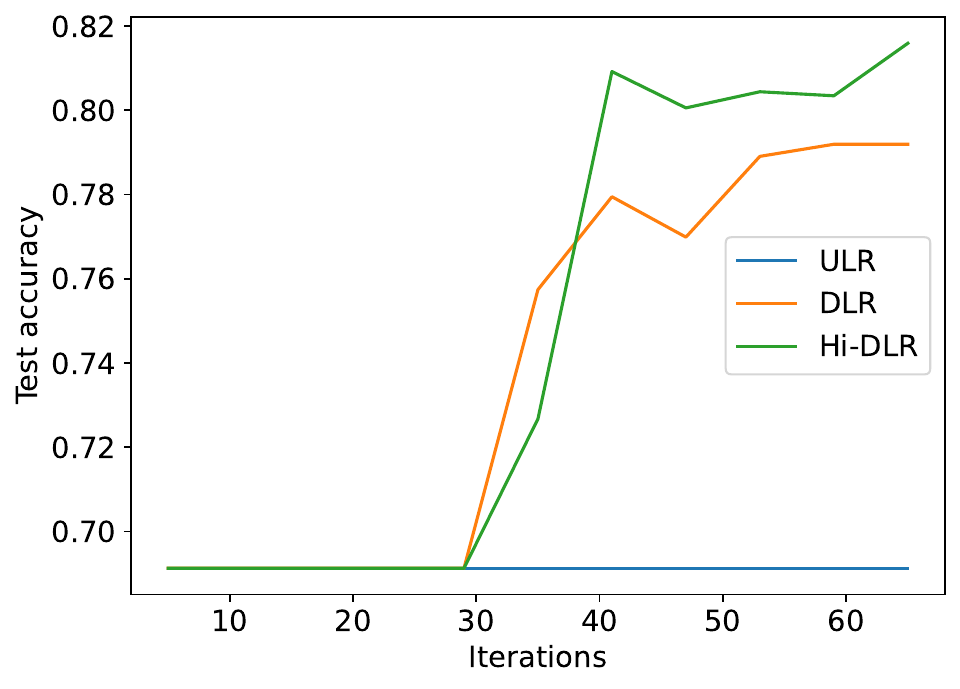}
\includegraphics[width=0.32\linewidth]{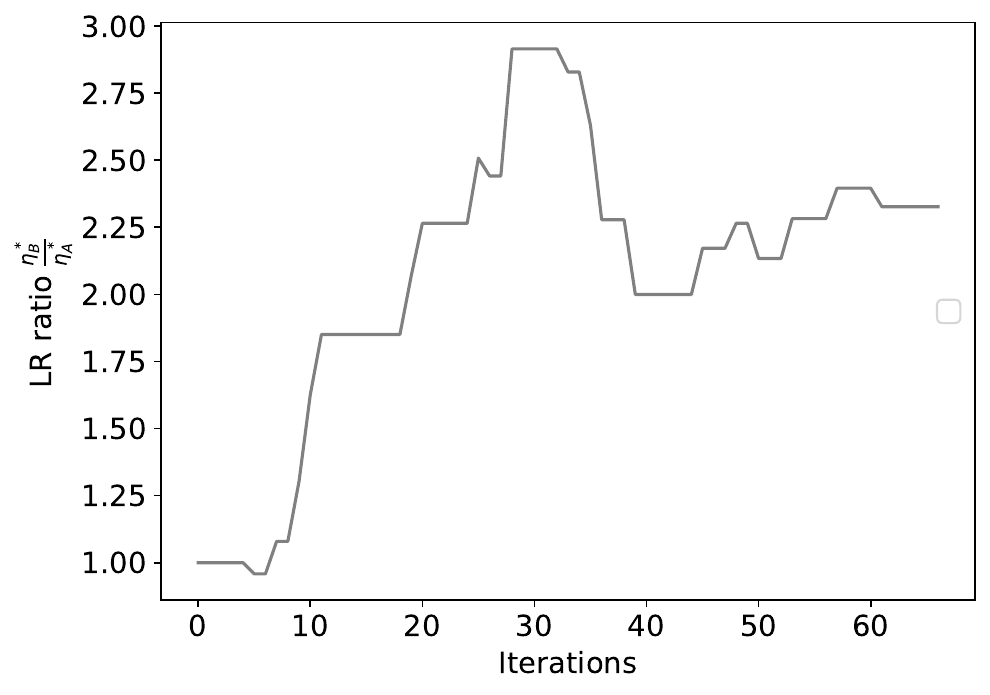}
    \caption{Hi-DLR outperforms manual ULR and DLR on multiple tasks optimized with LoRA (see experiment details in \ref{app:lora}). Left: synthetic regression. Middle \& right: text classification on the CoLA dataset in terms of accuracy and learning rate ratio $\eta_B/\eta_A$.
}
    \label{fig:dlr_illustration}
\end{figure}

For example, parameter-efficient fine-tuning (PEFT) methods including BitFit \cite{zaken2022bitfit}, low-rank adaptation (LoRA) \cite{hu2022lora}, prompt tuning (PT) \cite{lester2021power} and others (see a review in \cite{han2024parameter}) are special cases of the two-group DLR, because the majority of parameters is frozen and non-trainable (i.e. using a learning rate of 0) and a small portion of parameters is trained with a non-zero learning rate. These methods have shown strong performance in fine-tuning large vision and language models, including GPT, ViT, ResNet, etc. Furthermore, LoRA+ \cite{hayou2024lora+} has demontrated the benefit of applying two different learning rates to the two low-rank matrices in LoRA (see \Cref{fig:dlr_illustration} and the details in \Cref{sec:nlu}); DePT \cite{shi2023dept} has used different learning rates for the prompt encoder and low-rank matrices to improve performance. Another example is the layer-wise or block-wise learning rate. \cite{howard2018universal} proposed a depth-wise DLR where deeper layers use larger learning rate with $\eta_{(l)}=\eta\cdot 2.6^l$ and $l$ is the layer index. Similar ideas have been proposed not only in fine-tuning but also in pre-training \cite{you2019large,zheng2019blockwise,you2017large,singh2015layer,ginsburg2019stochastic,zhang2022dnn,sun2019fine,ioannou2023adalip}.

The examples above have shown that training with DLR may prevent us from getting a sub-optimal performance due to the under-training of some parts and over-training of the others. Despite the success of DLR in many areas, there are some challenges for its wider application which we introduce in the following.

\paragraph{Potential challenges in DLR.}
For $K$ parameter groups, DLR introduces $K$ learning rates as hyperparameters. This leads to the challenge of hyperparameter optimization, which can be prohibitively expensive,
especially when model size $D$ or number of learning rates $K$ is large.

One approach that reduces the number of effective hyperparameters in DLR, which is adopted by the aforementioned works, is to incorporate a heuristic structure among $\eta_{(k)}$ so as to reduce the degree of freedom: LoRA+ recommends to use a learning rate ratio $2^4$ between the two parameter groups, while the depth-wise learning rate uses a fixed ratio 2.6 to scale $\eta_{(l)}$, both reducing the degree of freedom in learning rates to effectively 1.

Nevertheless, such heuristic structure may fail to work in some cases. For example, the optimal learning rate ratio in LoRA+ for RoBERTa is $\approx 2^4$ but for LLAMA it reduces to $\approx 2^1\sim 2^2$; the ratio in depth-wise learning rate is expected to vary for different models and tasks.

To solve the hyperparameter optimization (HPO) of DLR, we consider an orthogonal approach that preserves the $K$ degrees of freedom and adaptively adjusts $\eta_{(k)}$ with minimal overhead, to be used in combination with any PEFT and adaptive optimizers.

\paragraph{Related work}
We briefly discuss some related work, including the HPO methods for learning rate that leverage the zeroth, first or second-order information. 

In practice, the manual grid search, as a zeroth-order HPO method is the workhorse in the case of ULR. However, the computational cost grows exponentially as $O(m^K)$, where $m$ is the searching range and $K$ is the number of hyperparameters, rendering it too expensive to be feasible for DLR ($K>1$) when the models are large.

Recent advances have proposed automatic (or parameter-free, or learning-rate-free) learning rate schedule, including but not limited to D-Adaptation \cite{defazio2023learning}, Prodigy \cite{mishchenko2023prodigy}, DoG \cite{ivgi2023dog}, and GeN \cite{bu2024automatic}. All these methods except GeN are the first-order HPO methods as they leverage the gradient information to design the learning rate via an estimation of $\eta_t=\frac{\|\w_0-\w_t\|}{\sqrt{\sum_{k=0}^t\|\g_k\|^2}}$. In contrast, GeN uniquely leverages the second-order Hessian information to enhance the design (see the formula in \eqref{eq:eta star}). Nevertheless, these methods in general are limited to the HPO of ULR instead of the DLR.

On the other hand, we note there are many learning rate techniques that can leverage Hessian information \cite{drori2020efficient, goujaud2022optimal,armijo1966minimization,bertsekas1997nonlinear}, but an extra and heavy overhead is incurred for a fine-grained search.

In this work, we extends GeN from ULR, so as to automatically solve the HPO of DLR in \Cref{alg:autoSGD}. We term our method as \textit{Hessian-informed DLR} (Hi-DLR) and highlight that its subcase Hi-ULR is equivalent to GeN. We have developed multiple efficient tricks in \Cref{sec:efficiency}, so that Hi-DLR remains state-of-the-art performance at minimum overhead, given any grouping of model parameters and any optimizer. 

\paragraph{Contribution.}
\begin{itemize}
    \item We introduce Hi-DLR in \eqref{eq:Hi-DLR} as an \textbf{automatic} HPO for differential learning rate. We highlight that Hi-DLR utilize the \textbf{second-order} Hessian information to enrich the pre-conditioning of any optimizer, so as to leverage the different loss curvature of different parameters through different learning rates.
    \item We design the highly \textbf{efficient} \Cref{alg:autoSGD} compute our Hi-DLR, with a novel diagonalization trick in \eqref{eq:lr parabola diag}, which significantly reduces the computation cost from $O(K^2)$ to $O(K)$. We note that the computation cost is further reduced to $O(1)$ with the infrequent update of learning rates.
    \item We demonstrate that Hi-DLR is \textbf{empirically strong} on various tasks like image/text classification, regression, multi-task learning, as well as PEFT.

\end{itemize}

\section{Hyperparameter optimization for differential learning rate}
\subsection{Notations}
\label{sec:notation}
We denote $\w$ as the parameters of a model, while $\w_t\in\R^D$ represents the iteration $t$ and $\w_{(k)}$ represents the $k$-th parameter group. We use $[\w_{(1)},\w_{(2)}]\in \R^{m+n}$ to concatenate two parameter groups in $\R^m$ and $\R^n$. The same notation follows for other variables including the mini-batch gradient $\g\in\R^D$, and we denote the learning rates $\bm\eta_{[K]}=[\eta_{(1)},...,\eta_{(K)}]\in\R^K$ for $K$ parameter groups. We denote the loss as $L(\w)$, its first-order derivative as $\G(\w):=\frac{\partial L(\w)}{\partial \w}$ and its second-order derivative as $\H(\w):=\frac{\partial^2 L(\w)}{\partial\w^2}$. We omit $t$ when it is obvious from the context.

\subsection{Hyperparameter optimization by next-loss minimization}
\label{sec:local motivation}

We study the HPO from a local perspective of the next-loss minimization, along any direction $\d\in\R^D$ by $\w_{t+1}=\w_t-\d$. Using the Taylor expansion to capture the loss curvature, we get
\begin{align}
&\min_{\d} L(\w_{t+1})
=\min_{\d} L(\w_t-\d)
\approx\min_{\d} L(\w_t)-\G_t^\top\d+\frac{1}{2}
\d^\top\H_t\d
\label{eq:taylor}
\end{align}
The minimizer $\d_t^*$ of \eqref{eq:taylor} is $\H_t^{-1}\G_t$, which leads to the Newton's method as $\w_{t+1}=\w_t-\d_t^*=\w_t-\H_t^{-1}\G_t$. 

However, $\H_t\in\R^{D\times D}$ is hard to compute for large-scale optimization, because of the complication in second-order differentiation and the prohibitive memory cost to store $\H_t$. In practice, $\H_t^{-1}\G_t$ is approximated by $\eta_t \g_t^\text{optim}\equiv \eta_t \P^{-1}\g_t$, i.e. the pre-conditioned gradient multiplied with a proper learning rate, and thus $\H^{-1}\approx \eta\P^{-1}=(\eta \I)\cdot \P^{-1}$. The majority of existing methods focus on merging the Hessian information into $\P^{-1}$. For example, Adam \cite{kingma2014adam}, AdamW \cite{loshchilov2017decoupled}, AdaGrad \cite{duchi2011adaptive}, AdaDelta \cite{zeiler2012adadelta}, 
RMSProp \cite{hinton2012neural} use the square root of diagonal Fisher information as $\P^{-1}$; AdaHessian \cite{yao2021adahessian} and Sophia  \cite{liu2023sophia} use the diagonal Hessian information or Gauss-Newton decomposition.

Orthogonal to these works, DLR (with $K$ parameter groups) extends $\eta\I$ to a $K$-dimensional diagonal matrix, up to permutation of elements, 
\begin{align*}
\H^{-1}\approx
\begin{pmatrix}
\eta_{(1)}\I&0&...&0
\\
0&\eta_{(2)}\I&...&0
\\
0&...&\ddots&0
\\
0&...&0&\eta_{(K)}\I
\end{pmatrix} \P^{-1}
\end{align*}
Consequently, DLR enriches the approximation to $\H^{-1}$ with a higher degree of freedom, which can be beneficial because the loss curvature can be very different for different parameters, as demonstrated by Figure 1-5 in \cite{ghorbani2019investigation}, Figure 1 and 3 in \cite{yao2020pyhessian}, Figure 1 and 6 in \cite{sankar2021deeper}, and Figure 1-2 in \cite{zhang2024transformers}. We visualize in \Cref{fig:2group} that, by grouping the parameters into biases and weights, the two groups have significantly different curvatures and prefer different learning rates.

\begin{figure}[!htb]
    \centering
    \includegraphics[width=0.25\textwidth]{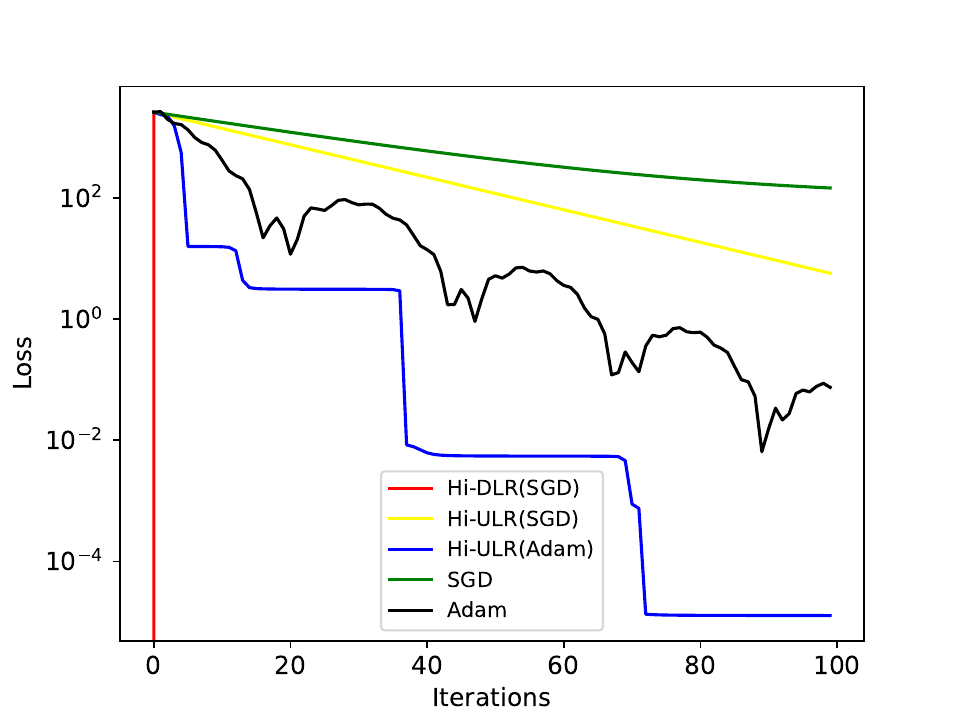}
    \includegraphics[width=0.24\textwidth]{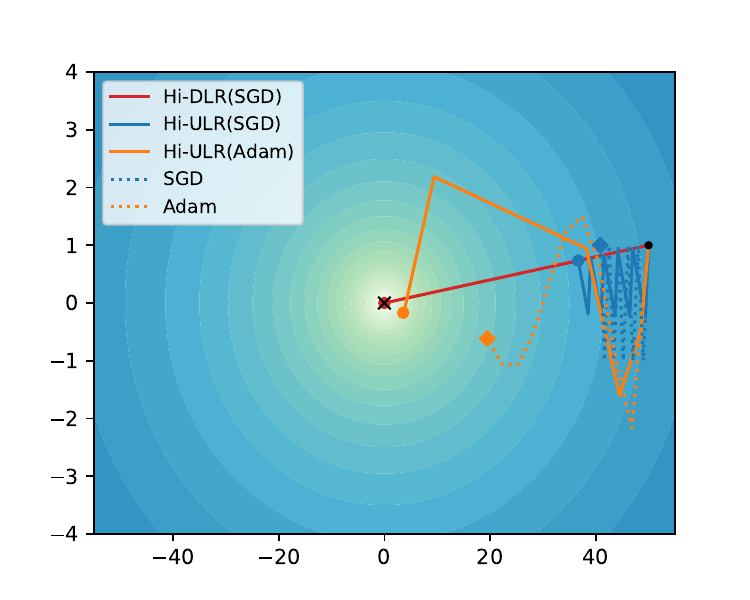}
    \includegraphics[width=0.25\textwidth]{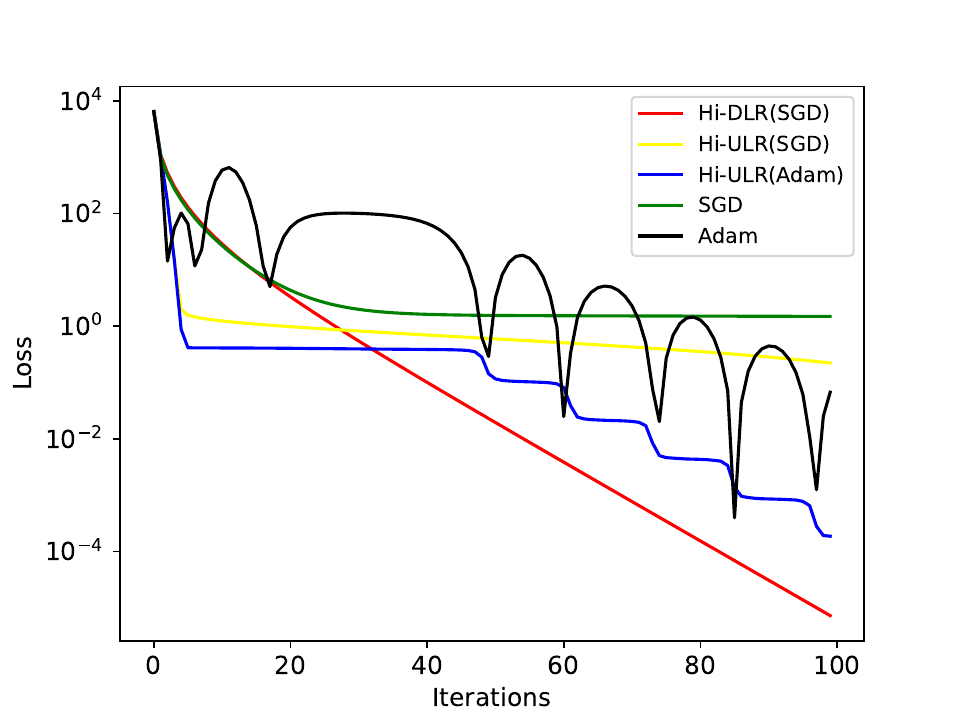}
    \includegraphics[width=0.24\textwidth]{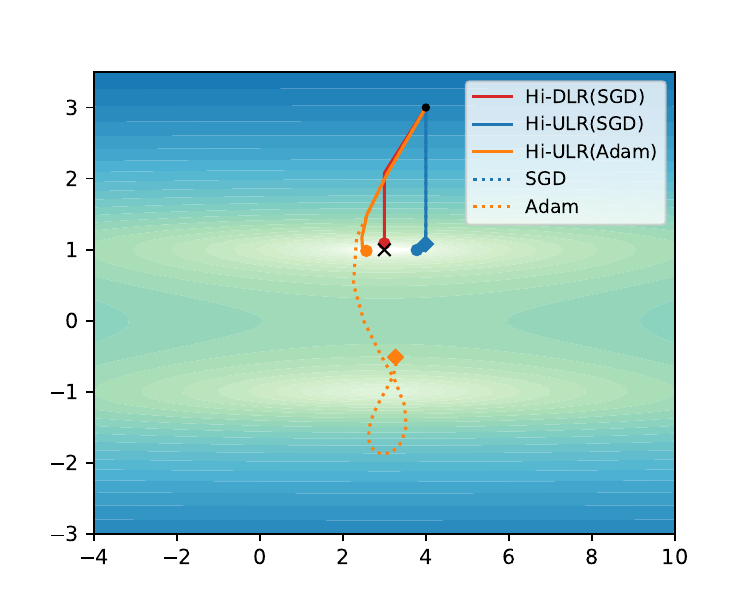}
    \caption{Optimizing over 2D test functions. The left two plots are the results of optimizing an ellipse function; the right two plots show the optimization on a function that is the sum of Beale and Rosenbrock. Hi-DLR is our method; Hi-ULR recovers GeN; the rest uses a manually selected learning rate. See experiment details in \Cref{app:syntheticDLR}.}
    \label{fig:synthetic}
\end{figure}

To put this into perspective, we test two functions in \Cref{fig:synthetic}: (1) the ellipse $L(w_0,w_1)=w_0^2+100w_1^2$, which is convex; (2) the sum of Beale and Rosenbrock functions, which is non-convex. We leave more details and explanation in \Cref{app:syntheticDLR}. We see that our Hi-DLR significantly accelerates the convergence\footnote{Specifically, for the ellipse function, we note that Hi-DLR reduces to the Newton's method, which is known to find the minimum in one iteration.} when compared to ULR. Additionally, although \eqref{eq:taylor} is a local minimization of one iteration, we have observed in \Cref{fig:synthetic} and our experiment sections that the advantage of DLR can be translated to $\min L(\w_T)$ throughout the training over multiple iterations.

\subsection{Optimal differential learning rates}
We now present Hi-DLR to solve the HPO in \eqref{eq:taylor} under $\d=\bm\eta_{[K]}\g^\text{optim}_{[K]}$, 
\begin{align}
&L(\w_{t+1})-L(\w_t)
= -\G^\top (\bm\eta_{[K]}\g_{[K]})
+\frac{1}{2}
(\bm\eta_{[K]}\g_{[K]})^\top
\H
(\bm\eta_{[K]}\g_{[K]})+o(|\bm\eta_{[K]}|^2)
\label{eq:temp taylor}
\\
\approx&-\bm\eta_{[K]}^\top
\underbrace{
\begin{pmatrix}
\G_{(1)}^\top\g_{(1)}\\...\\
\G_{(K)}^\top\g_{(K)}
\end{pmatrix}
}_{\b_*(\g^\text{optim}_{[K]})\in\R^{K}}+\frac{1}{2}\bm\eta_{[K]}^\top
\underbrace{
\begin{pmatrix}\g_{(1)}^\top\H_{(11)}\g_{(1)}&...&\g_{(1)}^\top\H_{(1K)}\g_{(K)}
\\...&...&...\\
\g_{(K)}^\top\H_{(K1)}\g_1&...&\g_{(K)}^\top\H_{(KK)}\g_{(K)}
\end{pmatrix}
}_{\A_*(\g^\text{optim}_{[K]})\in\R^{K\times K}}
\bm\eta_{[K]}
\label{eq:lr parabola}
\end{align}

\begin{figure}[!htb]
    \centering
\includegraphics[width=0.3\linewidth]{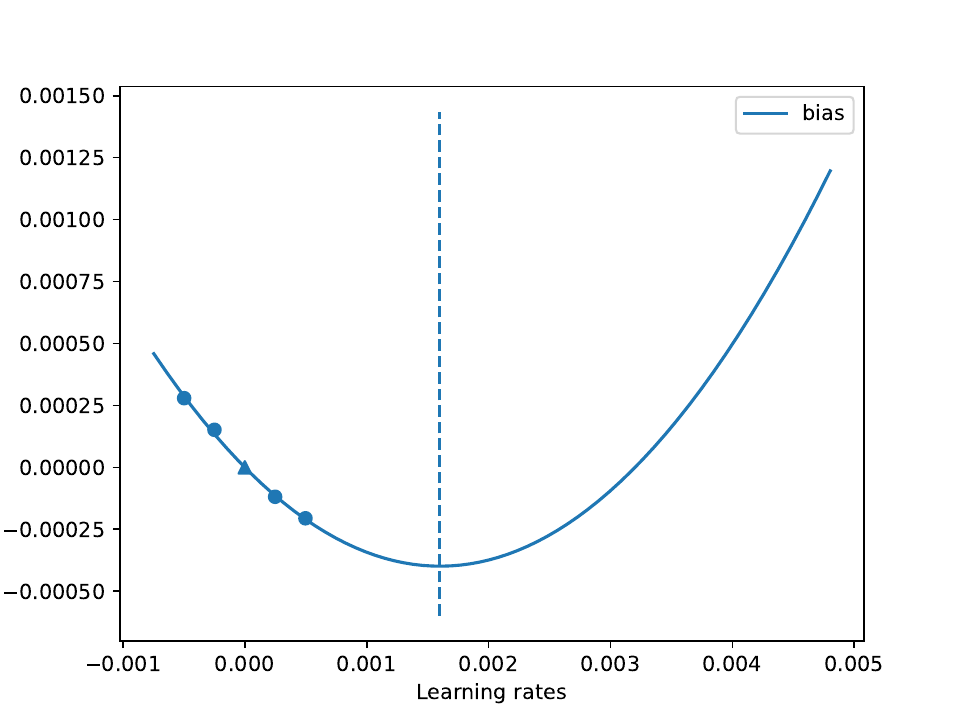}
\includegraphics[width=0.3\linewidth]{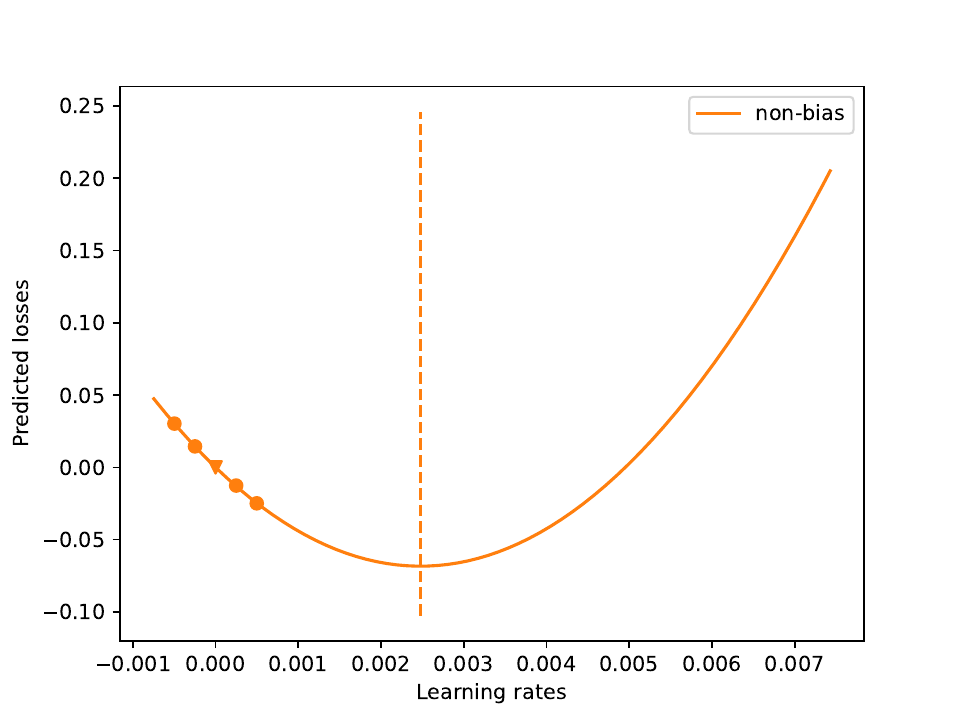}
\includegraphics[width=0.28\linewidth]{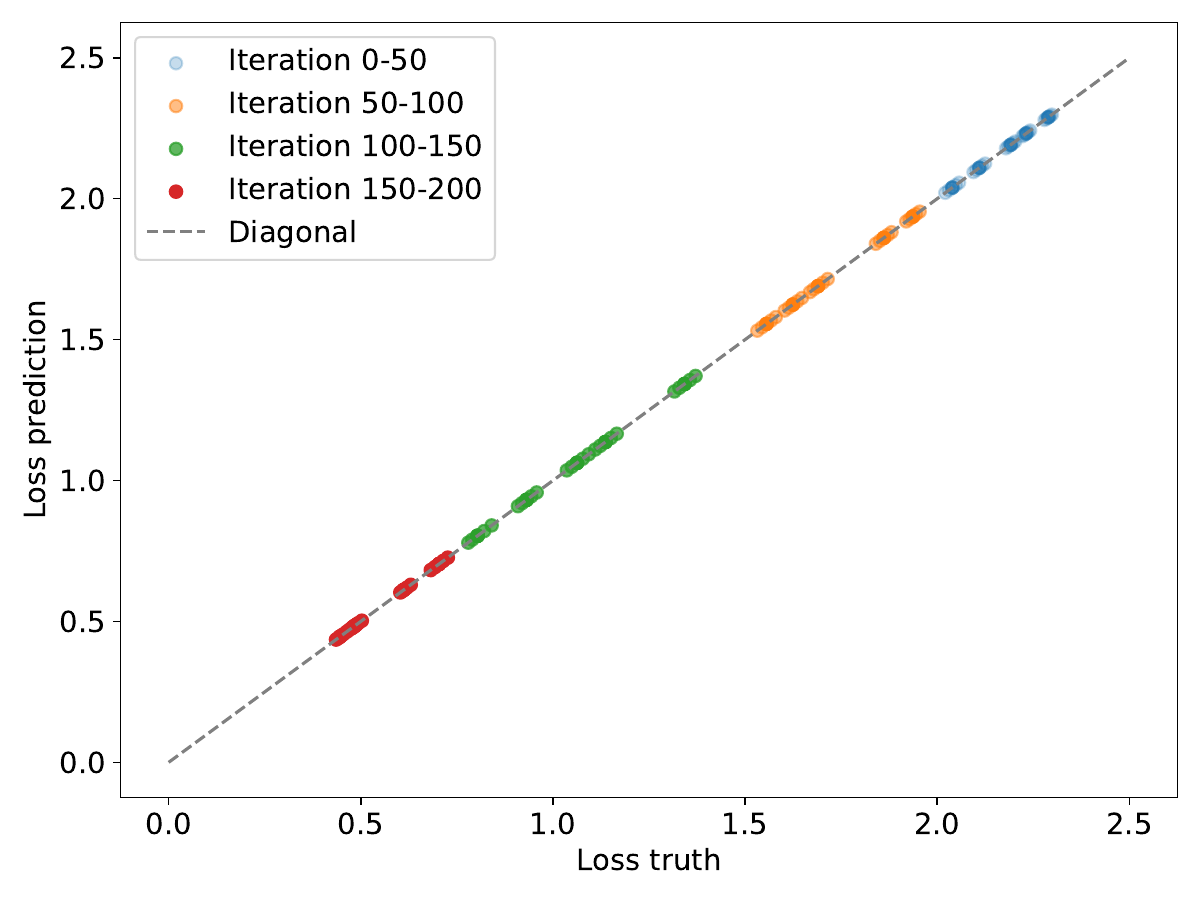}
    \caption{Second-order Taylor approximation in \eqref{eq:lr parabola diag} is sufficiently accurate. We visualize losses with two-group Hi-DLR (bias) under the settings in \Cref{sec:cv}. Left\&Middle: $L(\w_{(1)}-\xi_j\g_{(1)})$ and $L(\w_{(2)}-\xi_j\g_{(2)})$ in dots at iteration 200. Solid lines are the fitted quadratic functions, with minimizer marked by dashed vertical lines. Right: the loss truth is the left side of \eqref{eq:lr parabola diag} plus $L(\w)$, and the loss prediction is the right side of \eqref{eq:lr parabola diag} plus $L(\w)$.}
    \label{fig:2group}
\end{figure}

This approximation is sufficiently accurate when $\bm\eta_{[K]}$ is small (c.f. Figure 2 in \cite{bu2024automatic} when $K=1$; see also our \Cref{fig:2group}), because the error term $o(\eta^2)$ is very small for the commonly used learning rates.

If $\A_*$ and $\b_*$ are known and if $\A_*$ is positive definite, the quadratic function in \eqref{eq:lr parabola} admits a unique minimum at 
\begin{align}
\bm\eta_\text{Hi-DLR}=[\eta_{(1)}^*, ..., \eta_{(K)}^*]=\A_*^{-1}\b_*\in\R^K
\label{eq:Hi-DLR}
\end{align}
Notice that $\A_*$ and $\b_*$ can be defined on any $\g^\text{optim}$, hence Hi-DLR applies to any optimizer and the Hessian information is captured by both the pre-conditioning (through $\P^{-1}$ in $\g^\text{optim}$) and the learning rate (through $\A_*$ in $\bm\eta_\text{Hi-DLR}$). In what follows, we omit the superscript in $\g^\text{optim}$ for the simplicity of presentation.

\section{Computing Hi-DLR without additional back-propagation}
\label{sec:efficiency}
We propose \Cref{alg:autoSGD} to efficiently compute Hi-DLR, which requires the knowledge of $\A_*$ and $\b_*$ in \eqref{eq:lr parabola}, or equivalently $\G_{(k)}^\top\g_{(k)}$ and $\g_{(k)}^\top\H_{(kk)}\g_{(k)}$. Specifically, we demonstrate what, how, and when to derive these coefficients, thus reducing the computation overhead from $O(K^2)$ to $O(1)$ and allowing \Cref{alg:autoSGD} to be almost as fast as standard optimization. See our detailed complexity analysis in \Cref{complexity analysis}.

\begin{algorithm*}[!htb]
\caption{Generalized Newton's optimizers with multiple parameter groups}
\begin{algorithmic}[1]
\For{$t\in 1,\cdots,T$}
    \State Compute loss $L_0 = L(\w_t)$ by the forward pass
    \State Compute gradient $\g(\w_t)$ by back-propagation on $L_0$
    \State Modify gradient as $\g = \g^\text{optim}$ by AdamW, momentum SGD, etc.
    
    \If{$t \bmod \Phi == 0$}
        \State Let $\bm{v} = [-2, -1, 1, 2]^\top$,  $\bm{e}_i\in\mathbb{R}^K$ and $\bm{\eta}_{t-1}= [\eta_{(1),t-1}, \ldots, \eta_{(K),t-1}]^\top \in\mathbb{R}^K$
        \State Construct probing matrix:
        \[
            \bm E = 
            \begin{bmatrix}
                \bm{e}_1 \otimes \bm{v} \\
                \bm{e}_2 \otimes \bm{v} \\
                \vdots \\
                \bm{e}_K \otimes \bm{v} 
            \end{bmatrix}
            \cdot \operatorname{diag}(\bm{\eta}_{t-1})
        \]
        \For{$j \in 1, \ldots, 4K$}
            \State Set perturbation direction: $\bm{\hat\eta}^{(j)} = \bm{E}_{j,:}$
            \State Compute $L_j = L\left(\w_t - [\hat\eta_{(1)}^{(j)} \g_{(1)}, \ldots, \hat\eta_{(K)}^{(j)} \g_{(K)}]\right)$ by forward pass
        \EndFor
        \State Fit a quadratic curve from $\{\bm{\hat\eta}^{(j)}\} \rightarrow \{L_j - L_0\}$
        \State Derive $\G_{(k)}^\top \g_{(k)}$ and $\g_{(k)}^\top \H_{(kk)} \g_{(k)}$ in \eqref{eq:lr parabola diag}
        \State Derive optimal learning rates $\bm{\eta}^*$ by \eqref{eq:eta star} 
        \If{$\A_*\succ0, \b_*>0, \text{R2 score}>0.95$}
        \State Update the learning rate $\bm{\eta}_t=\gamma\bm{\eta}_{t-1}+(1-\gamma)\bm{\eta}^*$
        \EndIf
    \EndIf

    \State Update learning rate and parameters: 
    $\w_{t+1} = \w_t - [\eta_{(1),t} \g_{(1)}, \ldots, \eta_{(K),t} \g_{(K)}]$
\EndFor
\end{algorithmic}
\label{alg:autoSGD}
\end{algorithm*}

\paragraph{What to derive.}
$\A_*\in\R^{K\times K}$ contains $O(K^2)$ elements to be derived, which can be costly and hard-to-scale for large $K$ (say $K=40$ in CelebA), because we will use one forward pass to estimate each element. In practice, we simplify the multivariate quadratic function in \eqref{eq:lr parabola} by only deriving the diagonal of $\A_*$,
\begin{align}
&L(\w_{t+1})-L(\w_t)
\approx -\bm\eta_{[K]}^\top
\b_*+\frac{1}{2}\bm\eta_{[K]}^\top
\text{diag}(\A_*)\bm\eta_{[K]}
=\sum_k (\frac{1}{2}\eta_k^2 \g_{(k)}^\top\H_{(kk)}\g_{(k)}-\eta_k\G_{(k)}^\top\g_{(k)})
\label{eq:lr parabola diag}
\end{align}
which is minimized, if all $\g_{(k)}^\top\H_{(kk)}\g_{(k)}$ are positive, at
\begin{align}
\eta_k^*=\frac{\G_{(k)}^\top\g_{(k)}}{\g_{(k)}^\top\H_{(kk)}\g_{(k)}} \text{ for } k=1,...,K.
\label{eq:eta star}    
\end{align}
In summary, we derive $\text{diag}(\A_*)$ instead of the full $\A_*$, thus reducing the computation overhead from $O(K^2)$ to $O(K)$ with negligible accuracy degradation empirically.

\paragraph{How to derive.}
We adopt the back-propagation-free approach in \cite{bu2024automatic} to fit the quadratic function \eqref{eq:lr parabola diag}, without ever instantiating the computationally expensive $\G$ or $\H$. We solve a finite-sum problem:
$$\A_*,\b_*=\text{arg min}_{A,b} \sum_j |L(\w_t-\bm\xi_j\g_{[K]})-L(\w_t)+\bm\xi_j^\top \bm{b}-\frac{1}{2}\bm\xi_j^\top \bm{A}\bm\xi_j|^2$$
Note this is a multivariate problem with $2K$ variables and \Cref{alg:autoSGD} uses $4K$ different $\bm\xi_j\in\R^K$.

\paragraph{When to derive.}
We derive $\eta_k^*$ through $\A_*$ and $\b_*$ infrequently, say every $\Phi$ iterations following \cite{bu2024automatic}. This reduces the overhead from $O(K)$ to $O(1)$ if we set $\Phi=O(K)$. We do not update the learning rate if not all $\g_{(k)}^\top\H_k\g_{(k)}$ are positive, i.e. we use $\bm\eta_{[K]}$ from the previous iteration whenever \eqref{eq:lr parabola diag} is not convex in $\eta_k$.

\section{Experiments}
\label{sec:experiment PEFT}
In this section, we experiment with Hi-DLR for language modeling, image classification, multi-task learning, and regression. We leave the experiment details in \Cref{app:exp}.

\subsection{Performance on natural language understanding}\label{sec:nlu}

Low-Rank Adaptation (LoRA, \cite{hu2022lora}) is a popular PEFT method that adds two low-rank matrices, $\bm{A}$ and $\bm{B}$, to the pretrained weight matrix,
$$\w\to\w+\bm{B}\cdot\bm{A} \text{ where } \bm{A}\leftarrow\bm{A}-\eta_A \g_A, \bm{B}\leftarrow\bm{B}-\eta_B \g_B$$
and only trains the parameters in $\bm{B}$ and $\bm{A}$. Recent research has shown that freezing $\bm{A}$ (LoRA-FA, \cite{zhang2023lora}) or choosing different learning rates for $\bm{A}$ and $\bm{B}$ (LoRA+, \cite{hayou2024lora+}) can boost LoRA's performance. These variants can be viewed as applying DLR to the vanilla LoRA by using $\eta_B>\eta_A$.

\begin{table}[!htp]
  \centering
    \caption{Performance of RoBERTa-base model with different methods on GLUE datasets. 
    The best performance in PEFT is marked in bold.} 
  \footnotesize
  \begin{tabular}{l|r|ccccc} 
      &\begin{tabular}[c]{@{}c@{}}Trainable\\param\end{tabular} 
      & MNLI&SST-2 & MRPC & CoLA&QNLI\\
  \hline
    ULR (FMT)&125M&\textcolor{blue}{87.45}&\textcolor{blue}{94.38}&\textcolor{blue}{88.97}&\textcolor{blue}{80.82}&\textcolor{blue}{92.46}\\
    \hline
  ULR (LoRA) 
 &0.3M&85.01&93.81&75.49&69.13&\textbf{91.05}\\
  Hi-ULR (LoRA) &0.3M&82.49&93.35&83.58& 79.58&90.43\\
  Hi-DLR (LoRA) &0.3M&\textbf{\textcolor{red}{85.21}}&\textbf{\textcolor{red}{94.15}}&\textbf{\textcolor{red}{85.78}}& \textbf{\textcolor{red}{81.59}}&\textcolor{red}{90.48}
  \end{tabular}
\label{tab:NLU_results}
\end{table}

We fine-tune RoBERTa-base \cite{liu2019RoBERTa} model on five GLUE datasets \cite{wang2018glue} with LoRA. For Hi-DLR, we split the parameters into three groups: $\bm{A}$, $\bm{B}$ and \textit{head}. In \Cref{tab:NLU_results}, Hi-DLR outperforms Hi-ULR and ULR in PEFT on 4 out of 5 datasets. Experiment details can be found in \Cref{app:nlu}.

We notice that LoRA can underperform full model training (FMT) significantly on some datasets such as CoLA and MRPC. This phenomenon has also been witnessed in other models (see Table 1 of \cite{wang2024lora,wang2024lora2}). 
Additionally, Table 4 of \cite{bu2024automatic} shows that BitFit \cite{zaken2022bitfit}, another PEFT method can outperform LoRA on some GLUE datasets but not on others.

\subsection{Image classification}
\label{sec:cv}

We experiment on 5 image datasets for multi-class classification, in which we test 2-group Hi-DLR under full-model fine-tuning. We indicate one parameter group in the parenthesis in \Cref{tab:cv} (e.g. \textit{head}, \textit{bias}, and \textit{norm}), and treat the remaining parameters as the other group.

\begin{table}[!htb]
    \centering
        \caption{Test accuracy of ViT (optimized by AdamW) on image classification. We mark the best two results in bold for each dataset.}
    \begin{tabular}{c|c|c|c|c|c}
    Dataset&CIFAR10 & CIFAR100  & Food101 &GTSRB& SVHN\\\hline
        Reference&\multicolumn{2}{|c|}{\cite{krizhevsky2009learning}} & \cite{bossard2014food} &\cite{Houben-IJCNN-2013}& \cite{netzer2011reading}\\\hline
Hi-DLR (head) &98.80&93.03&\textbf{90.76}&\textbf{99.10}&96.73\\
Hi-DLR (bias) &\textbf{98.95}&\textbf{93.40}& \textbf{90.68}&\textbf{99.07}&96.80\\
Hi-DLR (norm) &98.86&\textbf{93.36}&90.45&99.06&96.82\\
\hline
Hi-ULR (GeN)&98.68&92.62&90.48&99.06&\textbf{97.14}\\
Prodigy&\textbf{98.92}&92.49&90.42&98.88&97.13\\
D-Adaptation&97.56&88.11&89.45&99.04&96.77 \\
\hline
Constant&97.49&89.23&88.44&98.54&96.65\\
Linear decay&98.48&92.60&90.54&98.74&97.08\\
Cosine decay&98.73&92.71&90.46&98.77&\textbf{97.16}    \end{tabular}
    \label{tab:cv}
\end{table}

Widely used ULR methods include heuristic learning rate schedulers (i.e. Constant \cite{raffel2020exploring}, Linear decay \cite{smith2015no} and Cosine decay \cite{loshchilov2016sgdr,radford2021learning}) as well as automatic optimizers like GeN, Prodigy and D-Adaptation. We compare Hi-DLR with these ULR methods and observe that Hi-DLR improves over the best ULR in all datasets except SVHN, since it takes our method some iterations to search the appropriate learning rates. 

\subsection{Multi-task learning}
We experiment on CelebA \cite{liu2015faceattributes}, a large-scale image dataset with 40 labels of face attributes and over 200k samples. This is a multi-label and multi-task problem, each label corresponding to one binary classification task. Hence we have 40 losses in total and will assign 40 learning rates to them. We use a pre-trained ResNet18 \cite{he2016deep} from \cite{rw2019timm} and only train the last layer, i.e. the classifier head. To be specific, the last layer has a shape $(512,40)$ and we group the parameters that connect the last hidden layer to each output neuron as one group with shape $(512,1)$, which corresponds to one task.

\begin{figure}[!htb]
    \centering
    \includegraphics[width=0.24\linewidth]{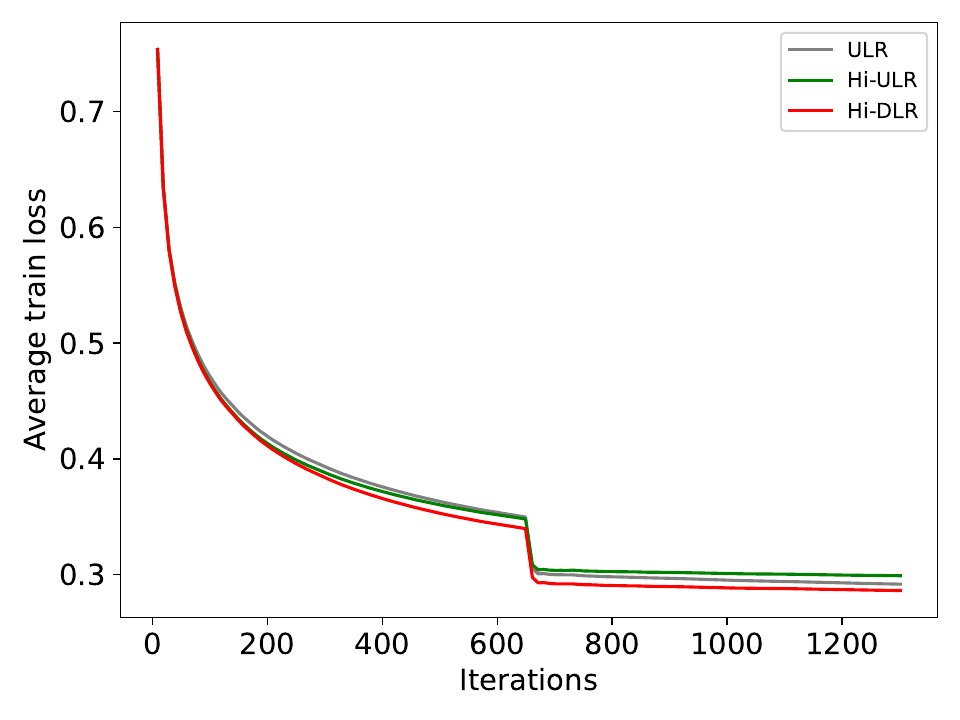}
    \includegraphics[width=0.24\linewidth]{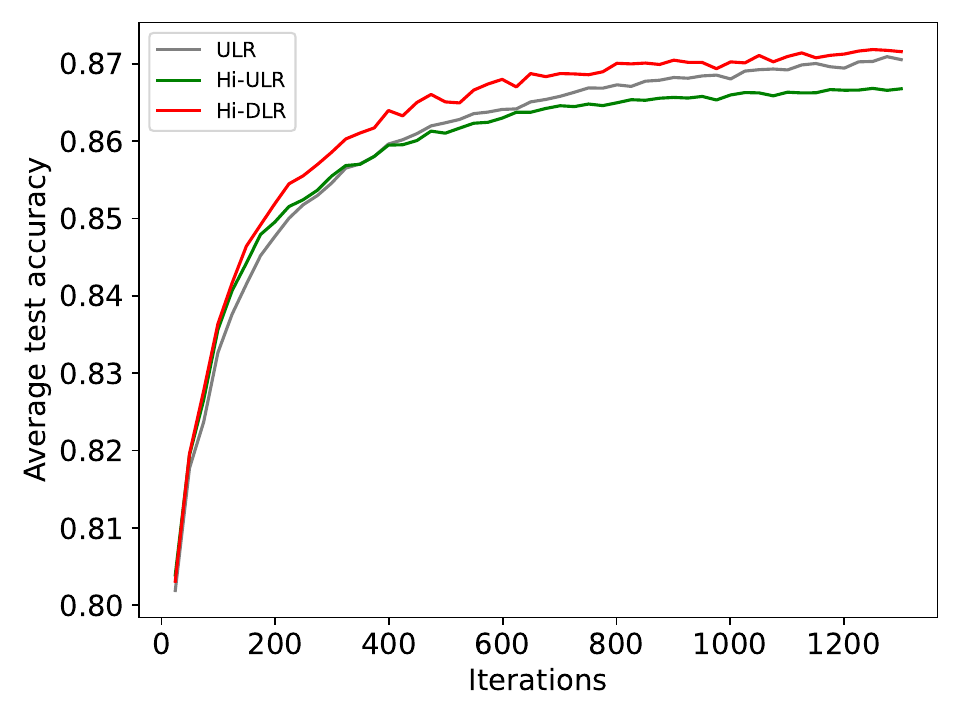}
    \includegraphics[width=0.24\linewidth]{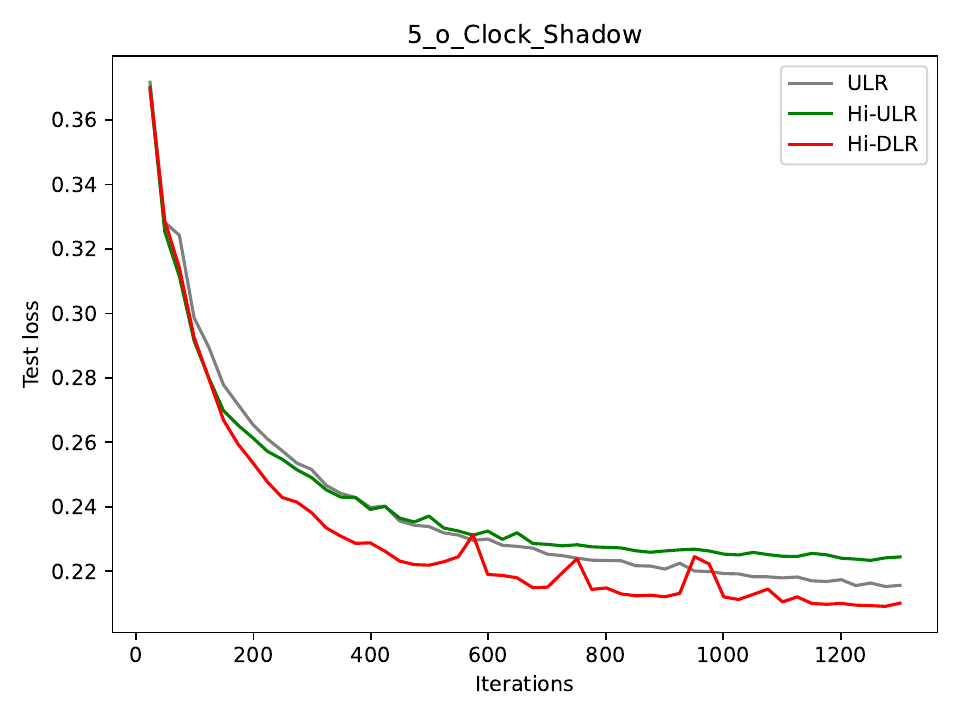}
    \includegraphics[width=0.24\linewidth]{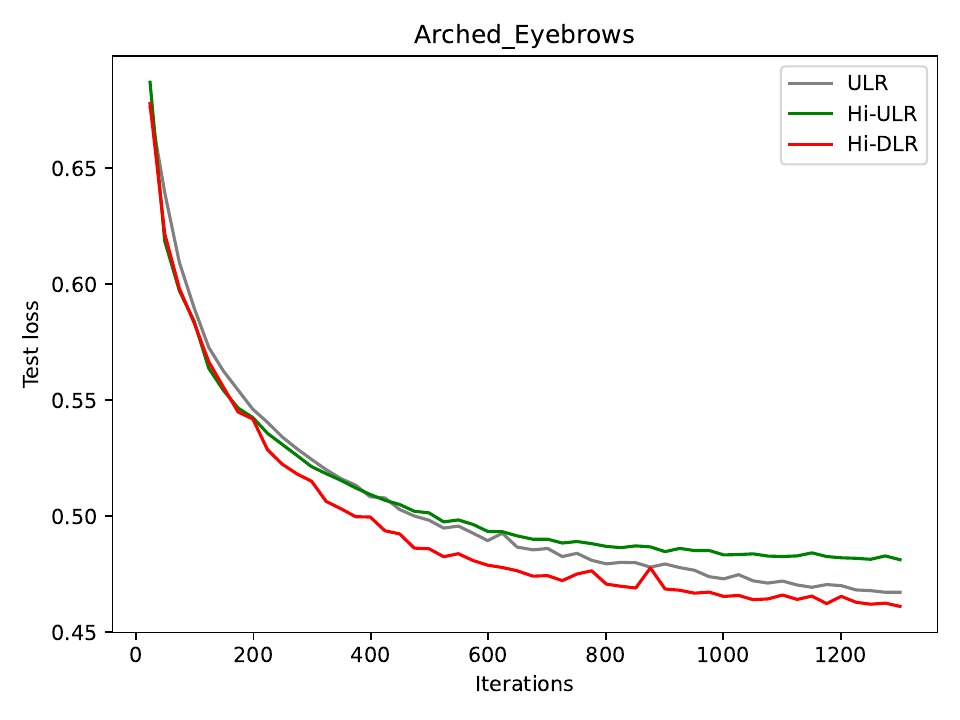}
    \vspace{-0.2cm}
    \caption{Fine-tuning results on CelebA. From left to right, the first panel shows the average train loss over 40 labels; the second panel shows their average test accuracy; the third and fourth panels are two individual test losses of two labels. See the results of all 40 tasks in \Cref{app:celeba}.}
    \label{fig:celeba}
\end{figure}

In \Cref{fig:celeba} (right two plots), we observe that the difficulty of learning different tasks can vary. Hence assigning different learning rates can improve both overall and individual convergence. 

\subsection{Interpretable regression with neural additive model (NAM)}
NAM \cite{agarwal2021neural, xu2023sparse} is a special neural network architecture, which has multiple sub-networks in parallel such that $g(\y)=\beta+\sum_{k=1}^{K} f_{j}\left(\bf X_{k}\right)$. Here $\y$ is the target variable, $g$ is the link function, $\bf {X_k}$ is the $k$-th feature of data, $\beta$ is the bias, and $f_k$ is the $k$-th sub-network. Each sub-network attends to a single feature separately so that the effect of each feature is interpretable.

Empirically, different features have various degrees of difficulty in learning, which requires different learning rates during training. We experiment on one synthetic data and the California housing dataset \cite{pace1997sparse}, as two regression tasks on tabular data. See experiment details in \Cref{app:nam}. 

\begin{figure}[!htb]
    \centering
    \includegraphics[width=0.32\linewidth]{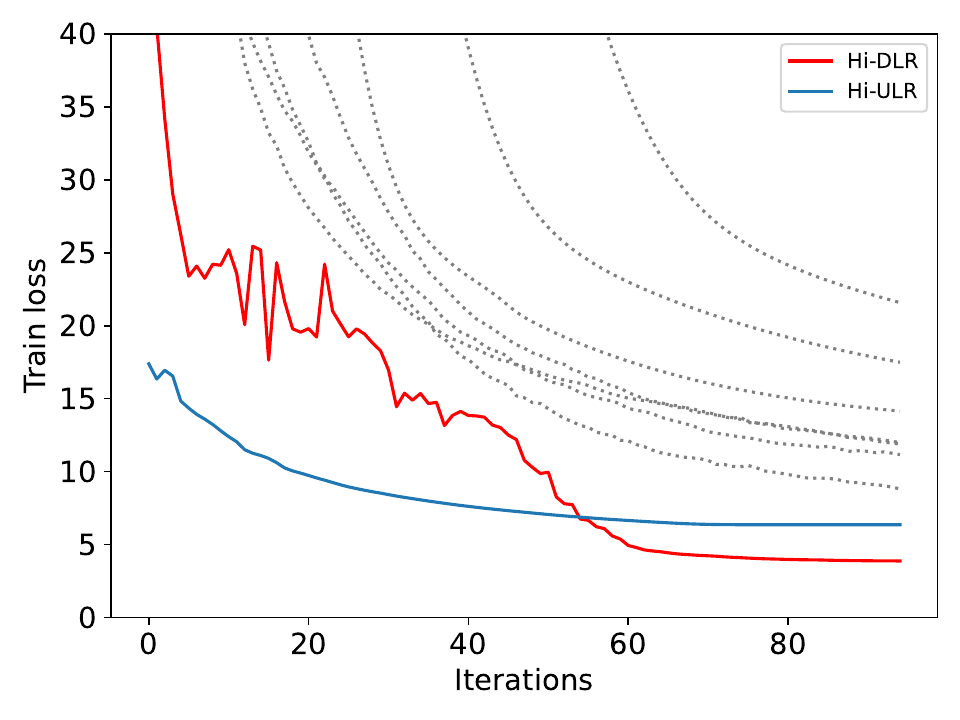}
    \includegraphics[width=0.32\linewidth]{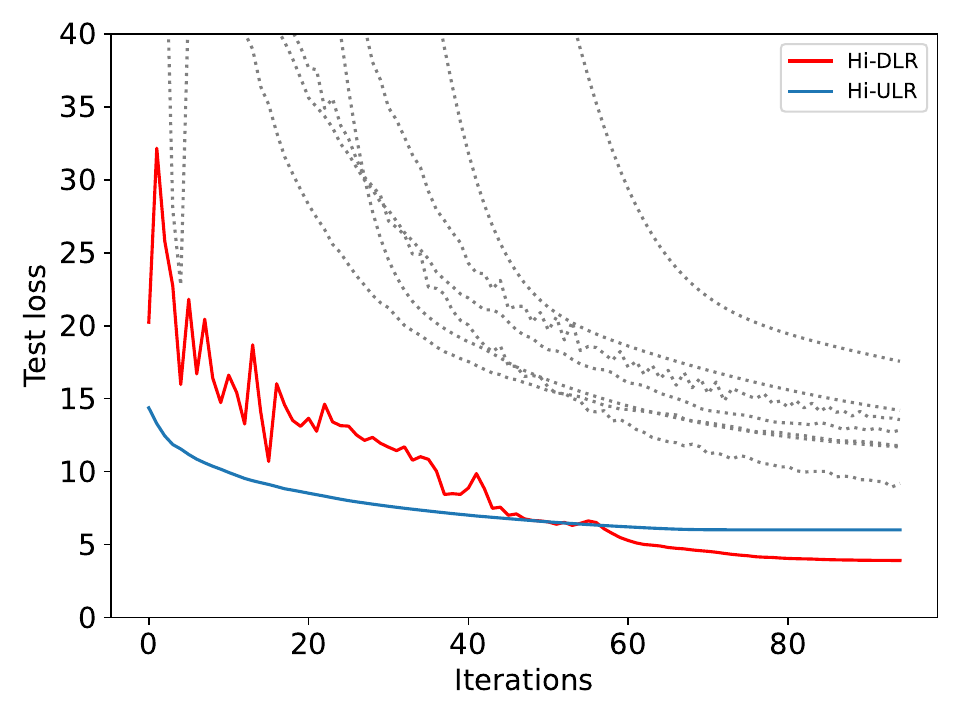}
    \includegraphics[width=0.32\linewidth]{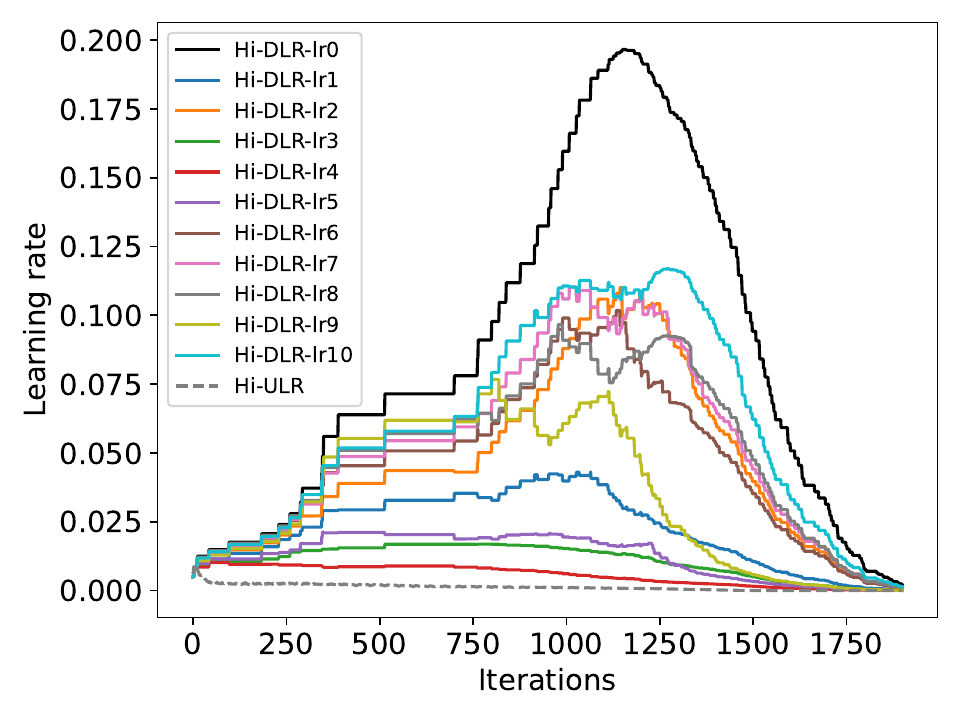}
    \includegraphics[width=0.32\linewidth]{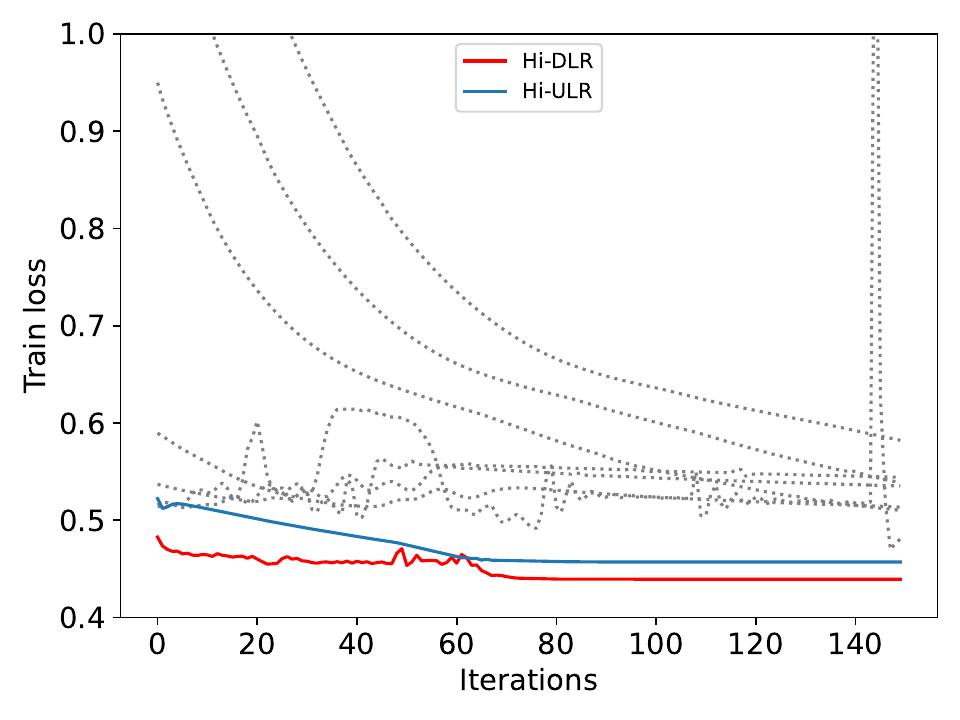}
    \includegraphics[width=0.32\linewidth]{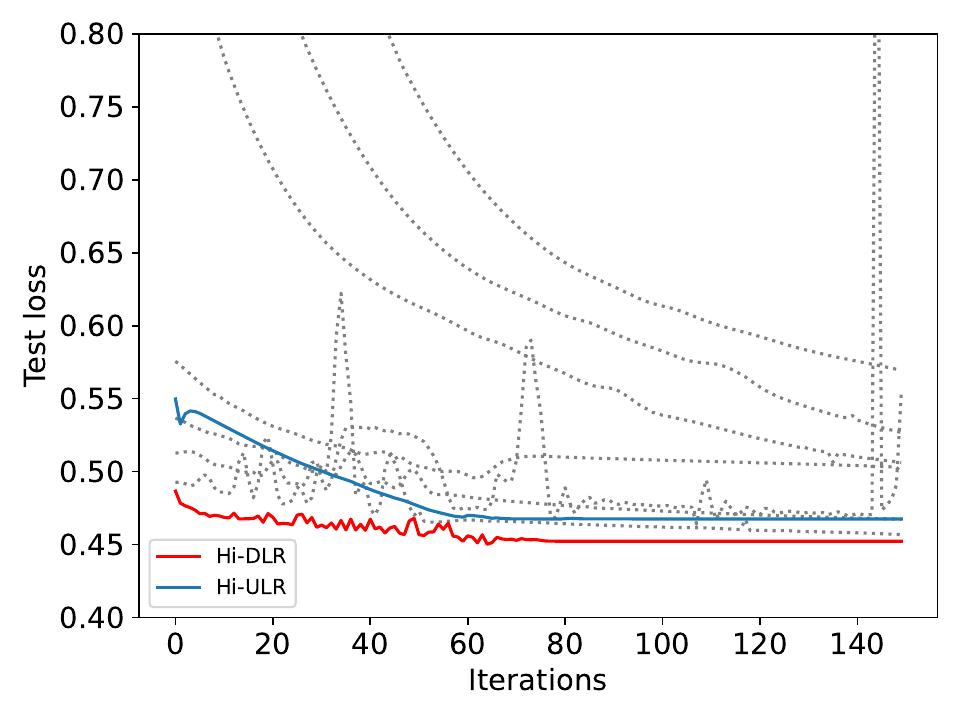}
    \includegraphics[width=0.32\linewidth]{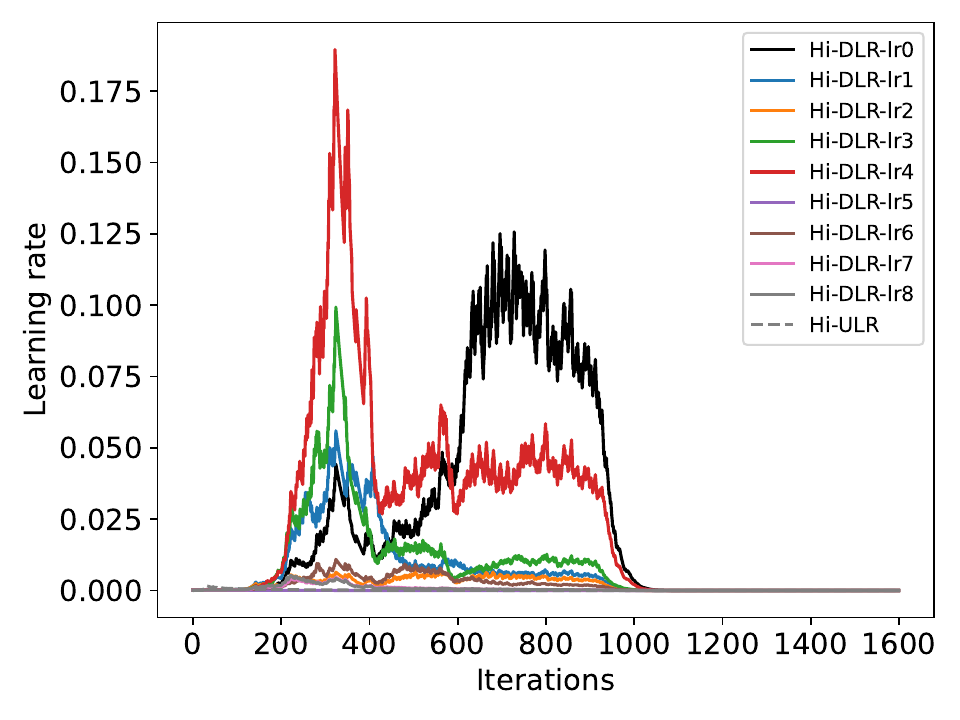}

    \vspace{-0.2cm}
    \caption{Loss and learning rate of NAM on two regression tasks. The first row is the synthetic dataset. The second row is the California Housing dataset. From left to right, the first two plots show the training losses, and test losses, where the \textbf{grey lines} are results trained with a list of manually picked learning rates, the blue curves correspond to Hi-ULR, and the red curves correspond to Hi-DLR; the last plot shows the learning rates for different groups.}
    \label{fig:NAM}
\end{figure}

We apply Hi-DLR to $f_k$ as follows: for $K$ sub-networks, we create $K+1$ parameter groups, with one for each $f_k$ and one for the bias $\beta$. The learning rates are shown in the right-most panel of \Cref{fig:NAM}. For Hi-DLR, the \textit{lr0} (black lines) is the learning rate for the bias. \textit{lr1}, \textit{lr2} $\cdots$ is the learning rate selected using Hessian information of parameter group $1,2,\cdots, K$.

In sum, the experiments in \Cref{fig:NAM} show that NAM with Hi-DLR converges significantly faster than manually selected learning rates or Hi-ULR. 

\section{Discussion}
In this work, we have demonstrated that different parameters have different loss curvatures and influences on the convergence, through the lens of Hi-DLR. We propose an efficient algorithm to adaptively compute Hi-DLR as an HPO solution, leading to faster convergence on various tasks. The success of DLR depends on the grouping of parameters: a sub-optimal grouping strategy might not lead to a good performance even with Hi-DLR. It remains an interesting future direction on how to efficiently find a good grouping strategy. Computation-wise, the training time of Hi-DLR increases linearly with the number of groups $K$ unless $\Phi$ also increases linearly, limiting its application to very large $K$ if the total number of iterations is small.

\newpage
\bibliographystyle{plain}
\bibliography{ref}

\newpage
\appendix
\section{Experiment details}\label{app:exp}
\subsection{Toy data for optimization}
\label{app:syntheticDLR}

To manually select the best learning rate, we grid search from $\{1e-5*10^{k/2}\}$ for $k=0,...,11$. The learning rate that gives the smallest loss after 100 iterations will be chosen.

\paragraph{Ellipse function} 
$$\text{Ellipse}(x,y)=x^2+100y^2$$.
We optimize from the initialization at $(x,y)=(50,1)$. The minimizer of the ellipse function is $(x,y)=(0,0)$.

\paragraph{Sum of Beale and Rosenbrock}
Beale is a convex function and Rosenbrock is a non-convex function. 
\begin{align*}
    \text{Beale}(x,y)&=(1.5-x+xy)^2+(2.25-x+xy^2)^2+(2.625-x+xy^3)^2\\
    \text{Rosenbrock}(x,y)&=100(y-x^2)^2+(1-x)^2
\end{align*}
The unique minimizer for Beale is $(3,0.5)$, for Rosenbrock is $(1,1)$.
The optimization problem is a sum of Beale and Rosenbrock:
$$L(x,y)=\text{Beale}(x,0.5)+\text{Rosenbrock}(1, y).$$
So the minimizer of this new $L$ is $(3,1)$. We optimize from the initialization at $(4,3)$.

\subsection{LoRA on natural language understanding}\label{app:lora}
\paragraph{Synthetic data}
Except for the GLUE benchmarks, we also experimented with a toy example in LoRA+ to better demonstrate DLR's power. The settings are the same as it is in Appendix C.1.1. of \cite{hayou2024lora+} except for $n$. We use $n=1000$ instead of $n=100$.

We train on 1000 iterations for each method and the plots start from the 50th epoch. 
For ULR, we grid search for the best learning rate based on the last test loss after 500 iterations. Assume $\eta_A$ and $\eta_B$ is the learning rate for $\bm A$ and $\bm B$ respectively. The search range for $\eta_A$ is ${10^{k}}$ for $k$ evenly searched from -4 to -3 for 20 points. The $\eta_B$'s search range starts from $k=-4$ to $k=-1$ for 20 points.

Finally, the selected ULR learning rates are $(\eta_A, \eta_B)=(1e-4, 1e-4)$. The best DLR learning rate are $(\eta_A, \eta_B)=(1e-4, 1e-1)$. 

\begin{figure}[!htb]
    \centering
    \includegraphics[width=0.4\linewidth]{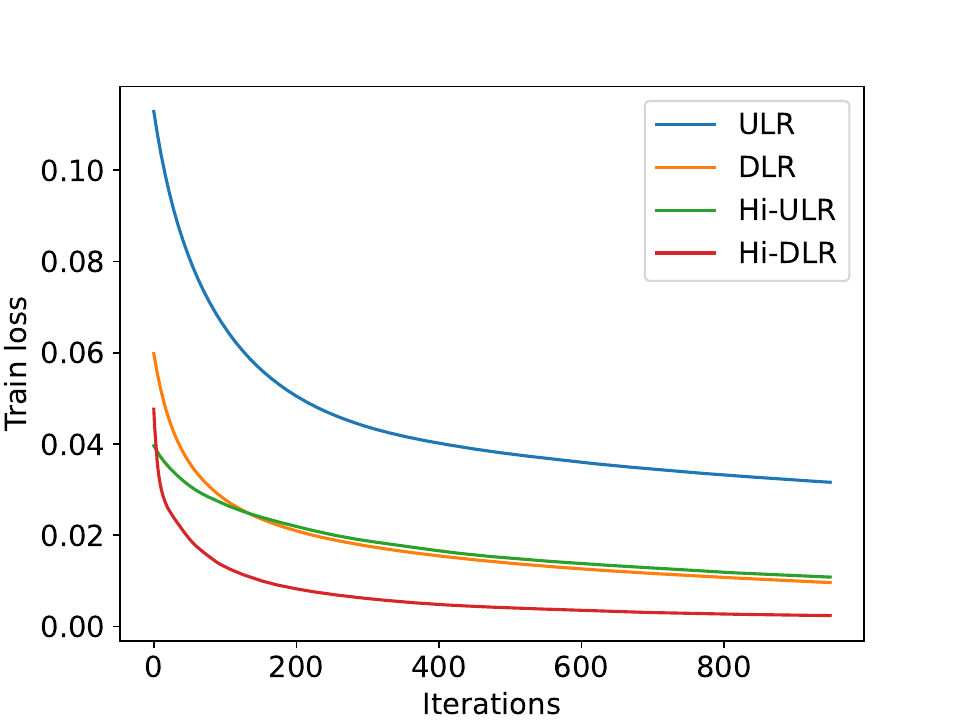}
\includegraphics[width=0.4\linewidth]{figs/loraplus_synthetic_test_loss.pdf}

    \caption{Hi-DLR outperforms manual ULR, DLR as well as Hi-ULR on a synthetic regression task optimized with LoRA.}

    \label{fig:lora+synthetic}
\end{figure}

\paragraph{NLU tasks}\label{app:nlu}
For NLU tasks, we use batch size 128 for all datasets. The evaluation metric is test accuracy. We use AdamW with a Cosine scheduler and warm-up ratio of 0.03. For every dataset, the full fine-tuning learning rates are 10 times smaller than their corresponding LoRA learning rate. The lazy frequency is selected based on batch size and data size.

\begin{table}[!htb]
\centering
\begin{tabular}{ccccc}
\hline
     & Data size & \begin{tabular}[c]{@{}c@{}}Initial learning rate \\ for FT\end{tabular} & \# of epochs     &$\Phi$             \\ \hline
MRPC & 3668       &         4e-5                                                             & 3&4                                    \\ \hline
SST2 & 67349        & 5e-5                                                                    & 3    &10                               \\ \hline
MNLI & 392702        & 5e-5    &1&10                                                                   \\ \hline
CoLA & 8551        & 4e-5                                                                    & 1       &1                 \\ \hline
QNLI & 104743        &    4e-5                                                                     &  3&  10                             \\
\hline
\end{tabular}
\caption{Hyper-parameters for GLUE training.}
\label{table:nluhyperparameter}
\end{table}
For hyper-parameters not mentioned here, we follow Table 9 of \cite{hu2022lora}.



   

\subsection{Prompt tuning on natural language understanding}\label{app:pt}
\begin{figure}[!htb]
    \centering
\includegraphics[width=0.325\linewidth]{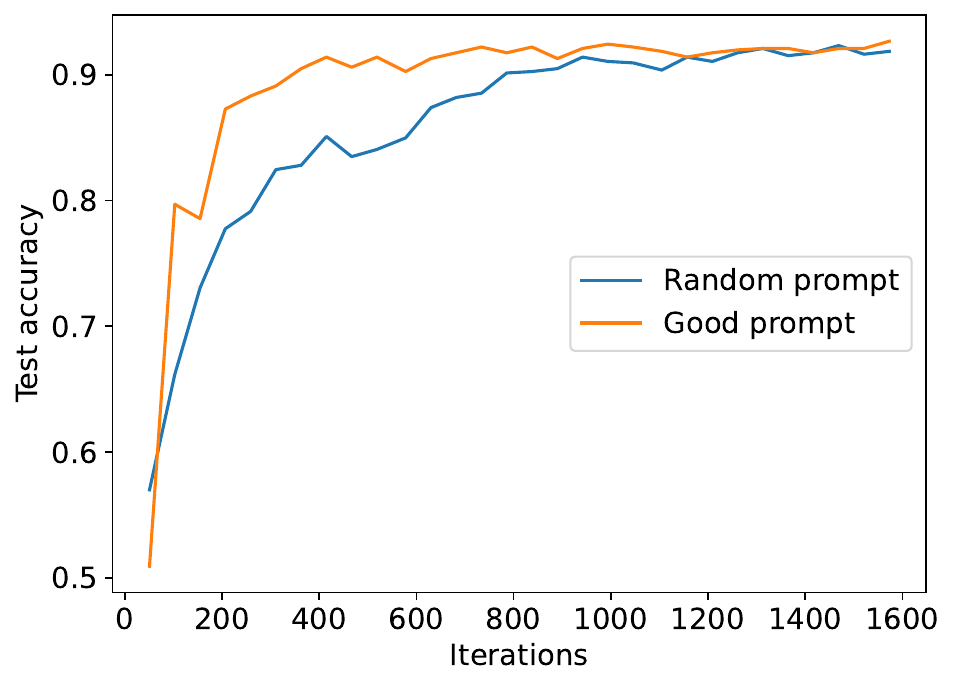}
\includegraphics[width=0.325\linewidth]{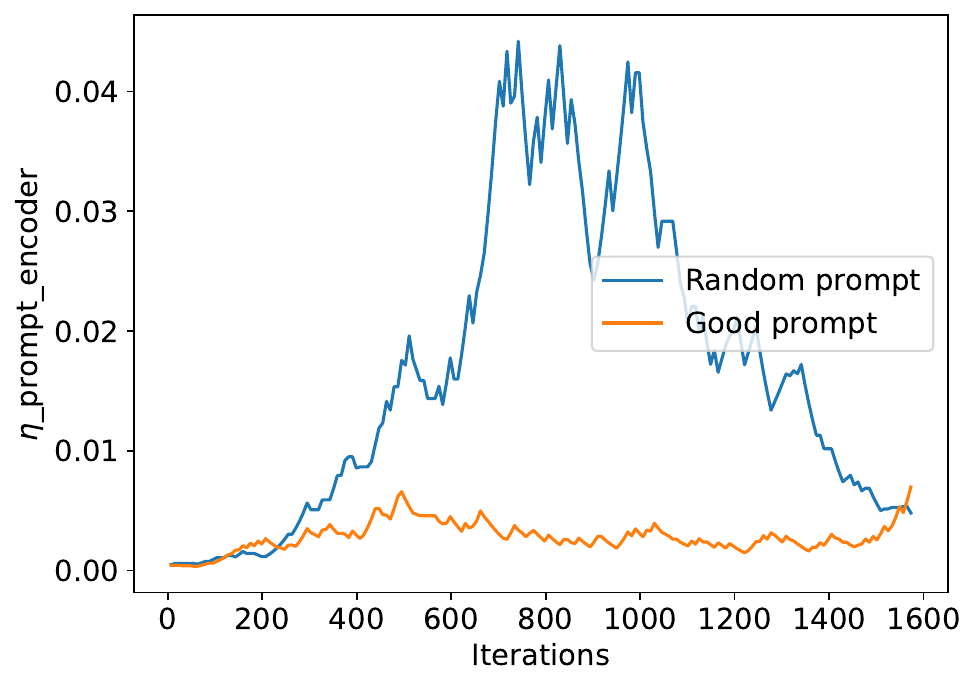}
\includegraphics[width=0.33\linewidth]{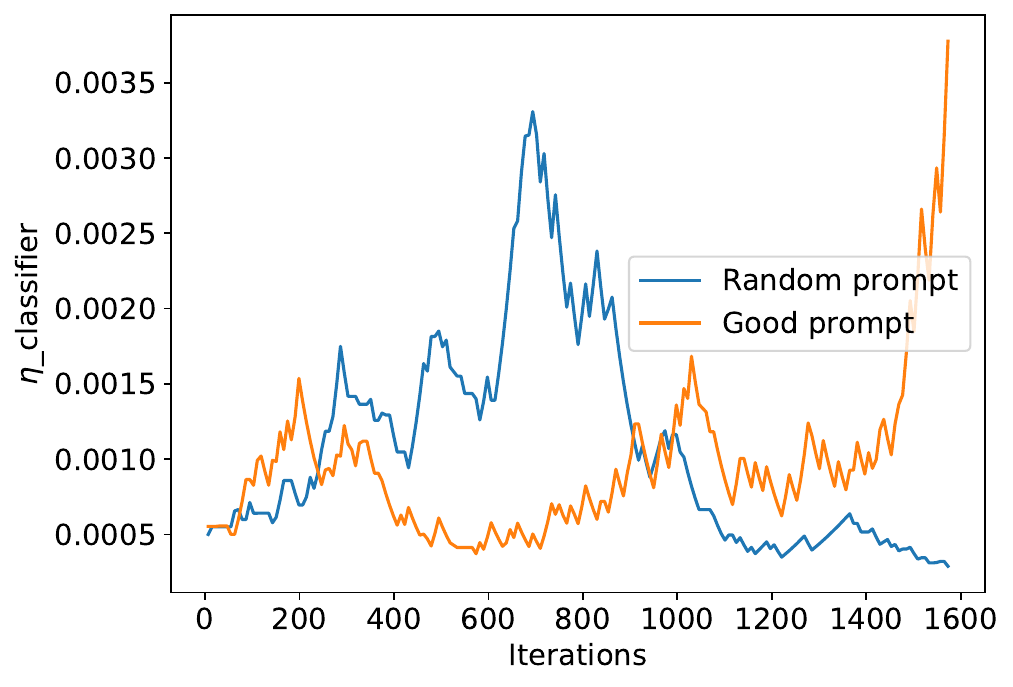}
    \caption{Applying Hi-DLR to prompt tuning for text classification on the SST2 dataset. We fine-tuned on two cases: a random initialized prompt (blue) and a good prompt (orange). In the right two figures, we plot the selected learning rates for the prompt encoder and the classifier.}
    \label{fig:pt_sst2}
\end{figure}

In \Cref{fig:pt_sst2}, a good prompt (orange) is \textit{"Predict if sentiment of this review is positive, negative or neutral"}, which gives a good initialization of the prompt encoder. 
In contrast, a random initialization (blue) requires a larger change from the original weights to achieve a comparable performance. The learning rates on the right two plots show Hi-DLR can adapt to different initializations: good initialization only needs small learning rate and vice versa for random initialization. 
We follow the default setting of prompt tuning from \href{https://huggingface.co/docs/peft/en/package_reference/prompt_tuning}{this tutorial}.

\subsection{GPT2}\label{app:nlg}
For GPT2, we experimented on the E2E dataset. The initial learning rate for full fine-tuning is 1e-4 while it is 1e-3 for PET. The sequence length is 128, the total batch size is 256 and the total validation batch size is 64. The total number of epochs for GPT2-small is 5, and for GPT2-medium and large is 3. 
The rest hyper-parameters are the same as in \cite{hu2022lora}. 

\section{Mixture-of-experts (MoE)}
We evaluate how different learning rate schemes affect a Mixture‐of‐Experts model on a challenging, non‐linear synthetic classification task. Leveraging our Hi-DLR framework, we assign differential learning rates to the gating network and expert modules -- refining routing decisions and mitigating early overfitting. 

We first synthesize 1000 training and 200 test points in $\mathbb{R}^2$, where labels are assigned by $\sin(x_1)+\cos(x_2)>0$ plus 10\% random label noise. The MoE comprises a two-layer gating network and six expert MLPs, each with a 64-unit hidden ReLU layer. We train four variants for 100 epochs (batch size 128) using Adam. We grid search the optimal learning rate for ULR (1e-3) and DLR (gate: 1e-3, experts: 5e-3). During training, we record training loss and test loss/accuracy. Compared to uniform and/or static learning rate schedules, Hi-DLR converges more rapidly and achieves higher test accuracy, demonstrating that dynamically adapting group‐wise learning rates can significantly improve learning in noisy, highly non‐linear MoE architectures.

\begin{figure}[!htb]
    \centering
    \includegraphics[width=0.49\linewidth]{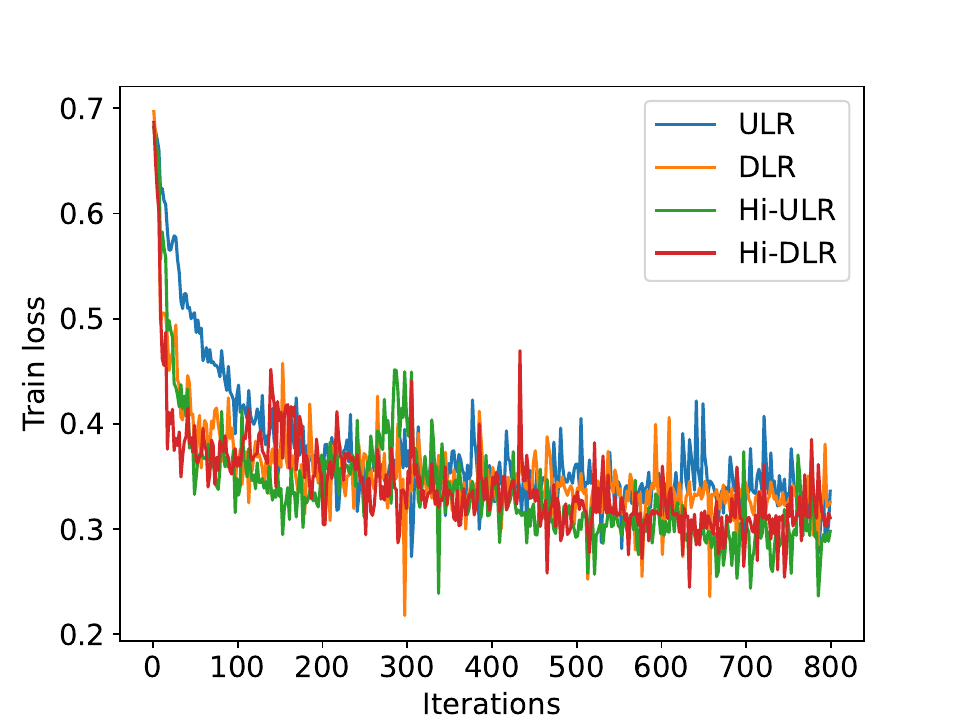}
    \includegraphics[width=0.49\linewidth]{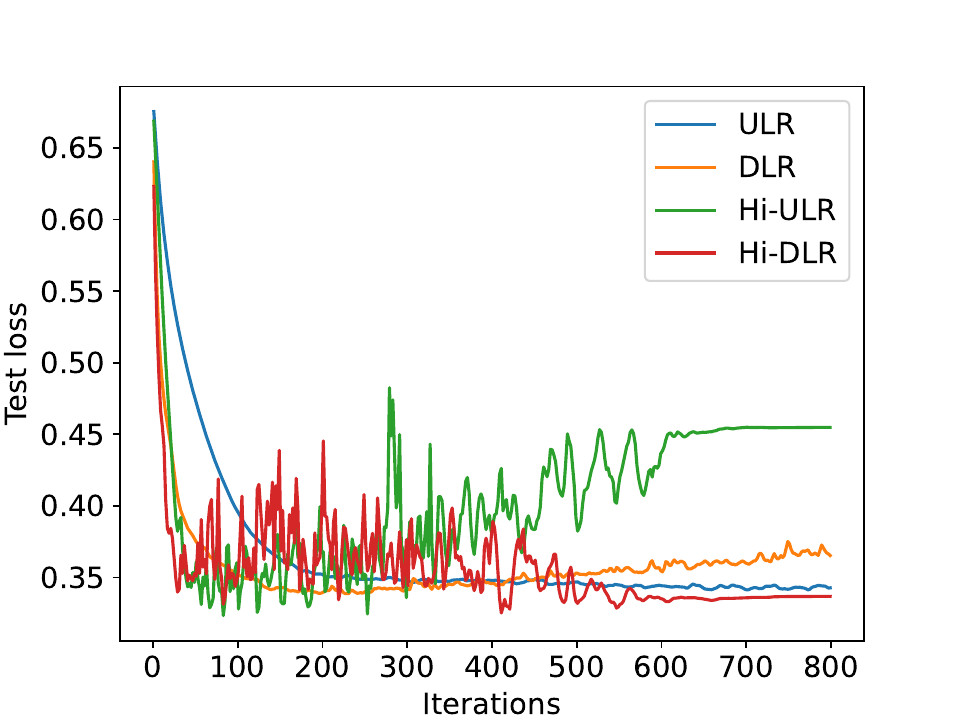}
    \includegraphics[width=0.49\linewidth]{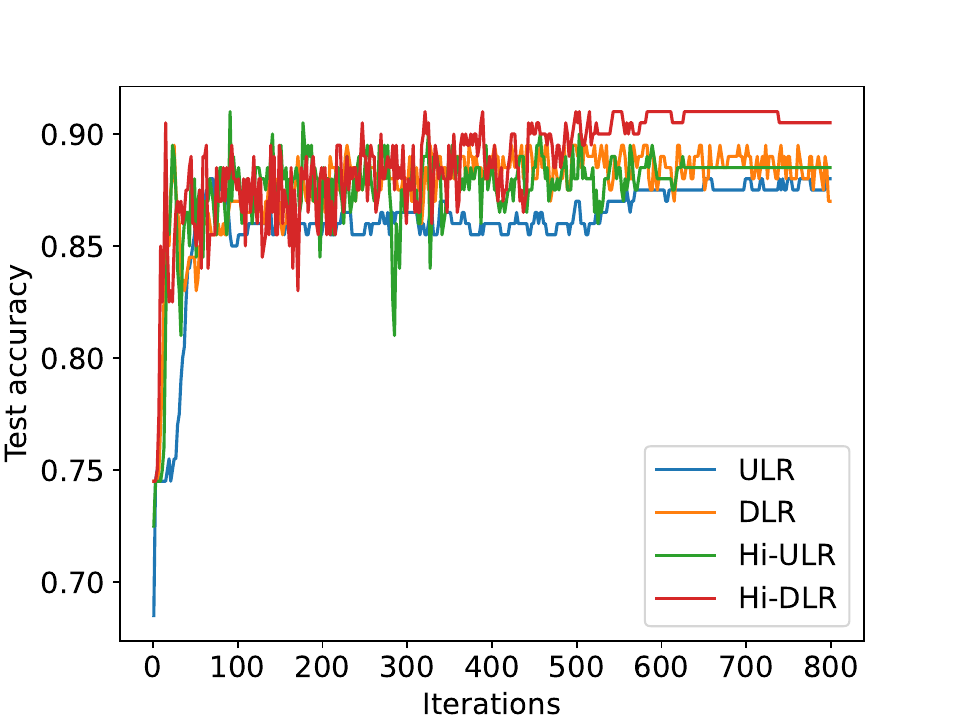}
    \includegraphics[width=0.49\linewidth]{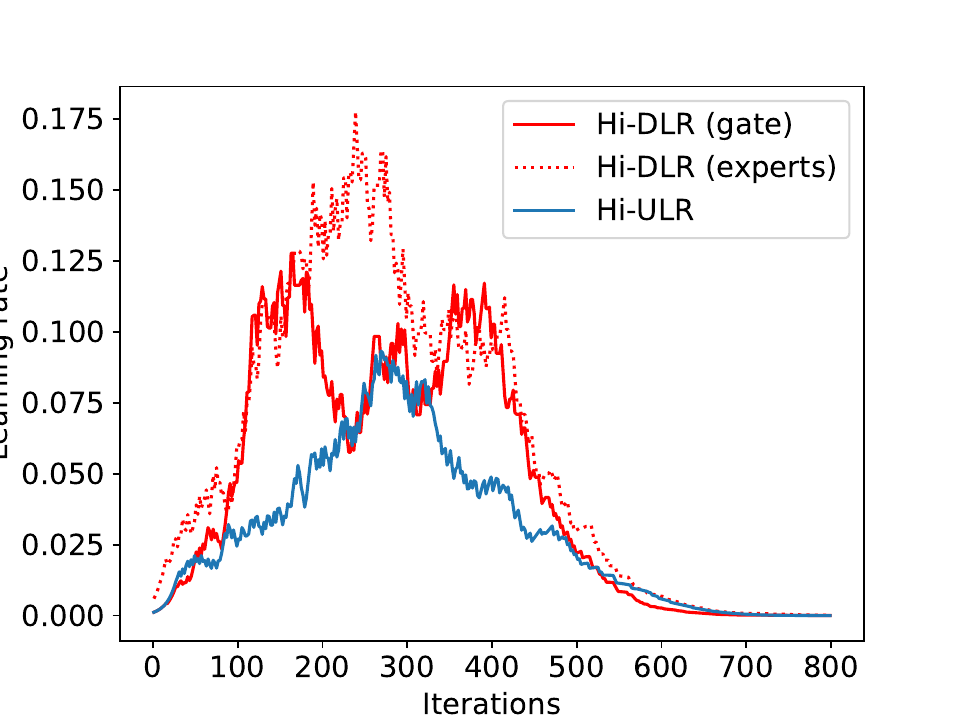}
    \caption{Training and evaluation of a six-expert MoE on a noisy, non-linear 2D classification task under four learning-rate schemes.}
    \label{fig:moe}
\end{figure}

\subsection{ViT classification}
We use the pre-trained \texttt{ViT-base-patch16-224} which can be can be loaded
from \texttt{timm} library. This model has been trained on ImageNet following \cite{dosovitskiy2020image}. We resize all images to 224x224 and normalize the pixel
values to [-1,1]. We use AdamW optimizer with the default
hyperparameters in Pytorch, except the learning rates. For methods that are not ours, we follow the learning rate settings in \cite{bu2024automatic}. For Hi-DLR, we use initial learning rate 1e-4, which is the same as Hi-ULR (GeN). We use batch size 500 across datasets with $\Phi=4$.

\subsection{Multi-task learning on CelebA}\label{app:celeba}

Each result is trained on 2 epochs with a training batch size of 500, optimized by a standard AdamW optimizer. No data augmentation is used. For ULR, we use a fixed learning rate of 1e-3. For Hi-ULR and Hi-DLR, we use an initial learning rate 1e-3 and $\Phi=10$.
\clearpage
\begin{figure}[!htb]
    \centering
    \includegraphics[width=.85\linewidth]{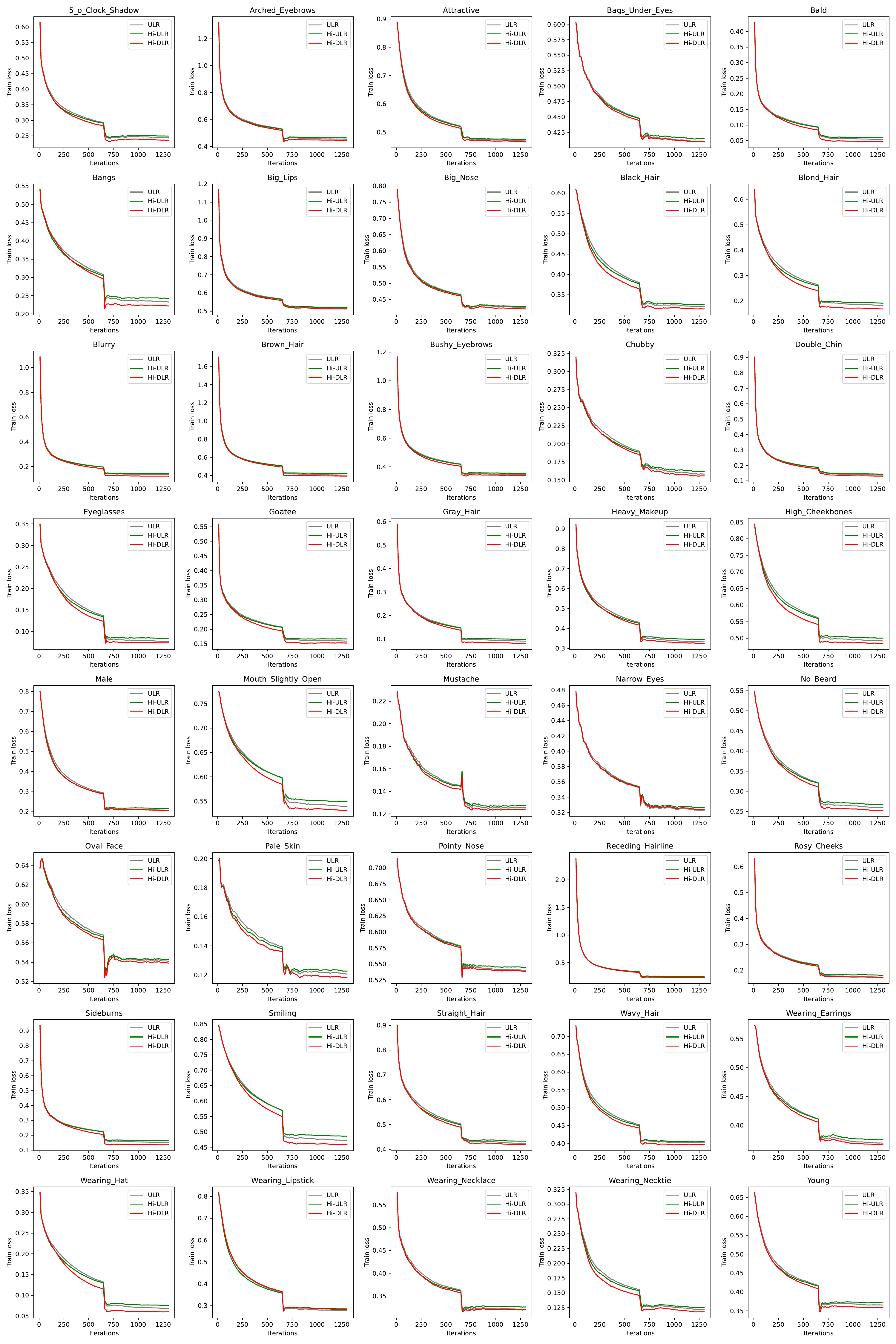}
    \caption{Individual train loss for 40 different labels of fine-tuning CelebA.}
    \label{fig:celebA_all_train_loss}
\end{figure}

\clearpage
\begin{figure}[!htb]
    \centering
    \includegraphics[width=.85\linewidth]{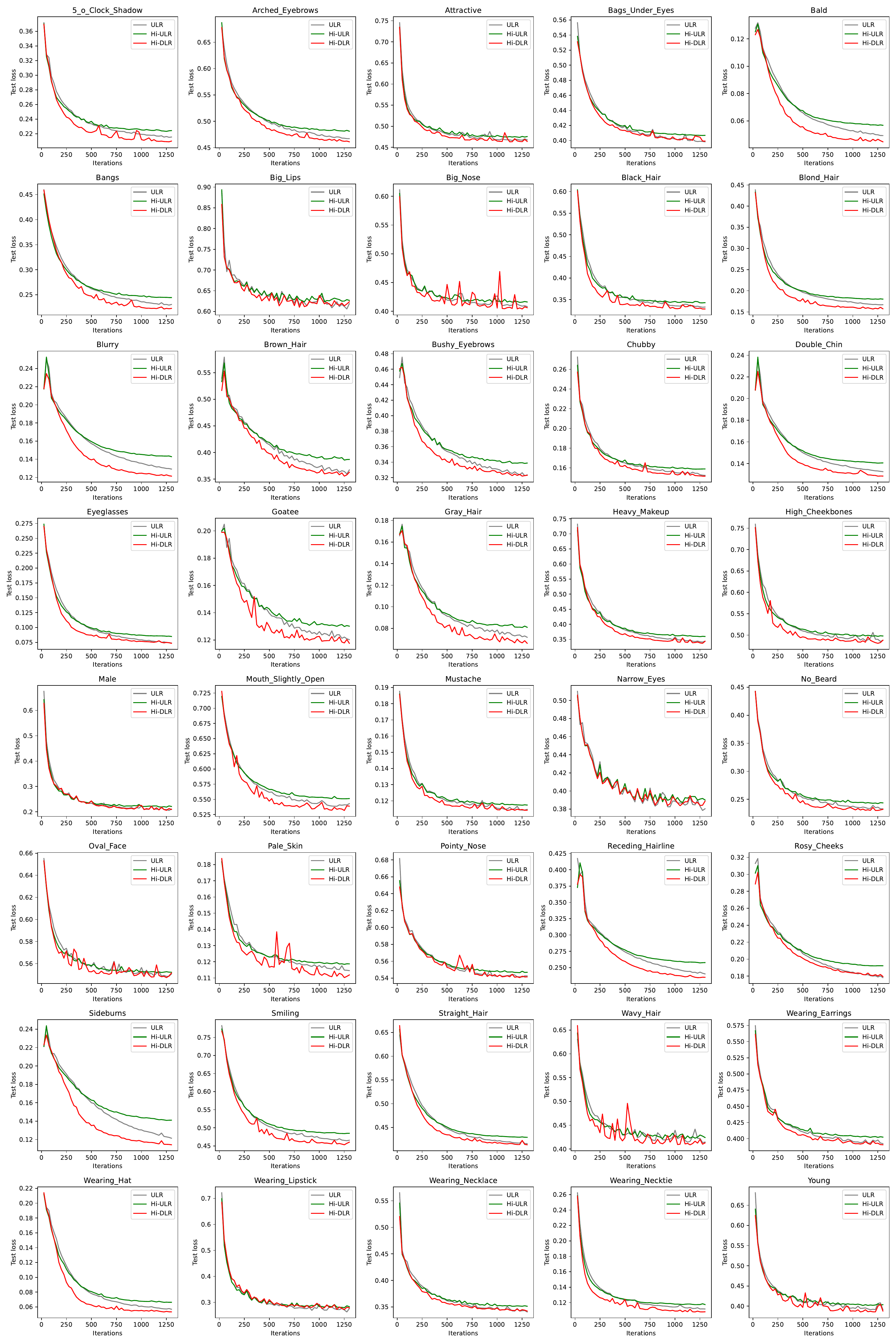}
    \caption{Individual test loss for 40 different labels of fine-tuning CelebA.}
    \label{fig:celebA_all_test_loss}
\end{figure}
\clearpage
\begin{figure}[!htb]
    \centering
    \includegraphics[width=.24\linewidth]{figs/celebA_avr_train_loss.pdf}
    \includegraphics[width=.24\linewidth]{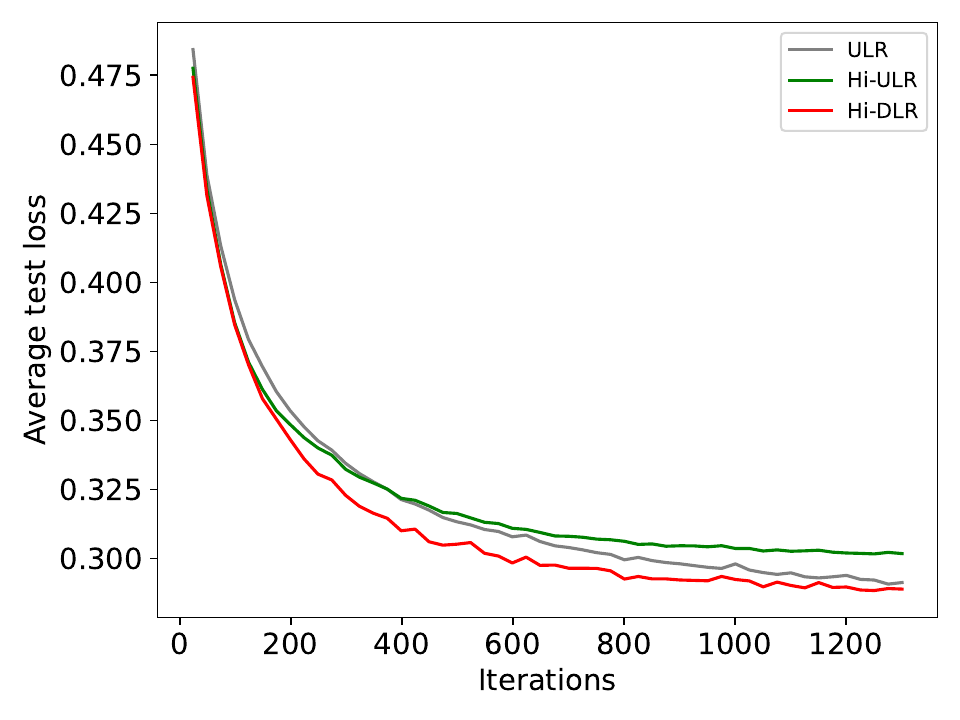}
    \includegraphics[width=.24\linewidth]{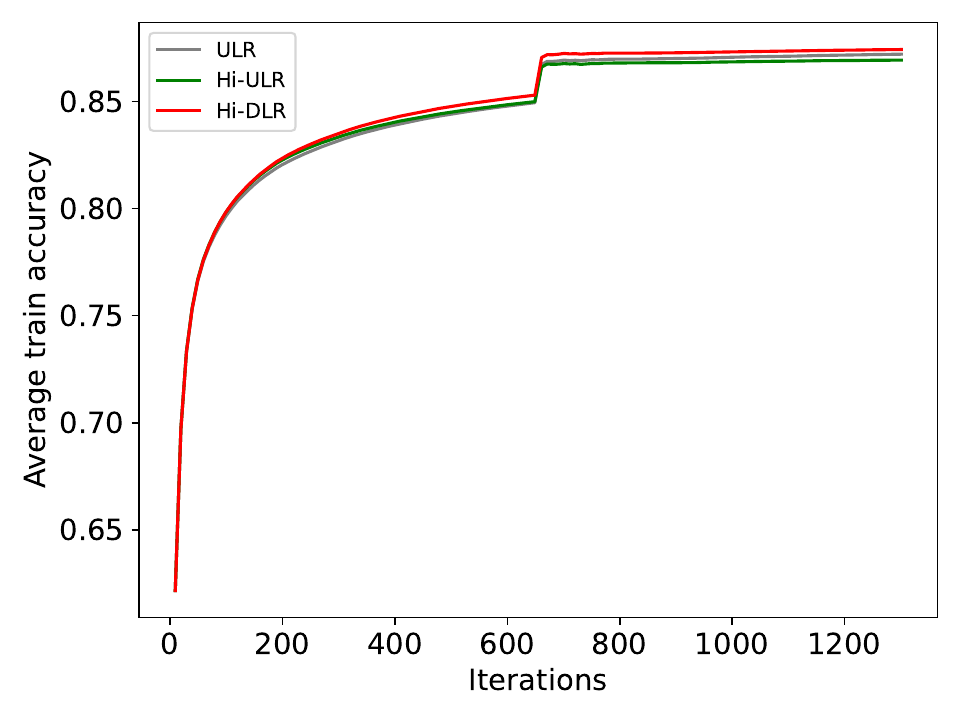}
    \includegraphics[width=.24\linewidth]{figs/celebA_avr_test_acc.pdf}
    \caption{Average performance of fine-tuning CelebA over 40 labels.}
    \label{fig:celebA_avr_performace}
\end{figure}

\subsection{Interpretable regression with NAM}\label{app:nam}
\begin{figure}[!htb]
    \centering
    \includegraphics[width=0.32\linewidth]{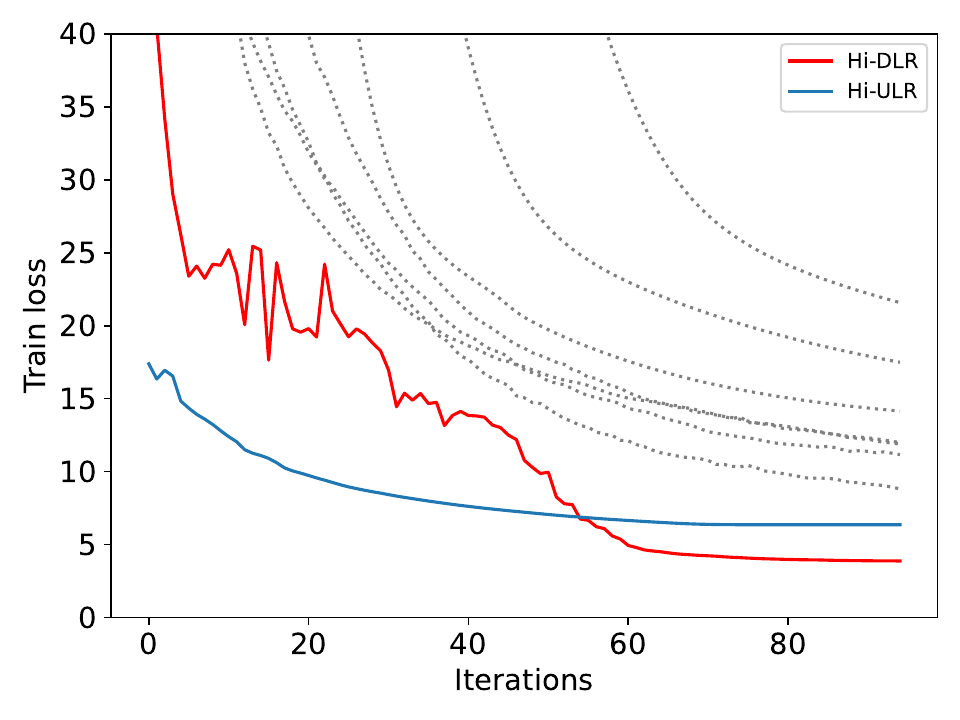}
    \includegraphics[width=0.32\linewidth]{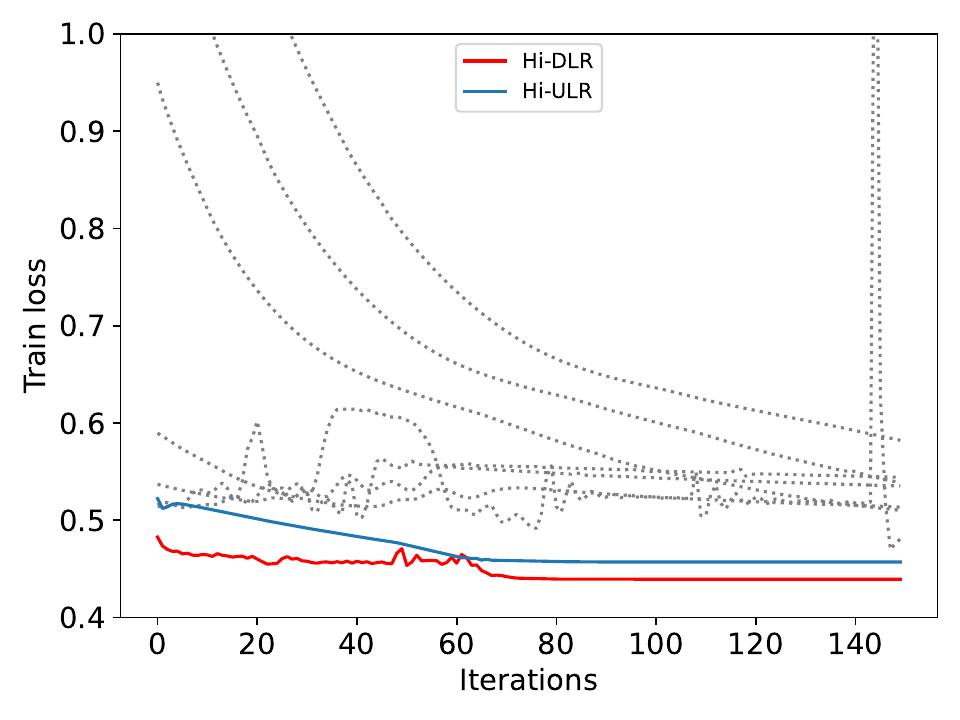}

    \caption{Train Losses of NAM on two regression tasks described in \Cref{fig:NAM}.}
    \label{fig:NAM_train}
\end{figure}
\begin{figure}[!htb]
    \centering
    \includegraphics[width=.88\linewidth]{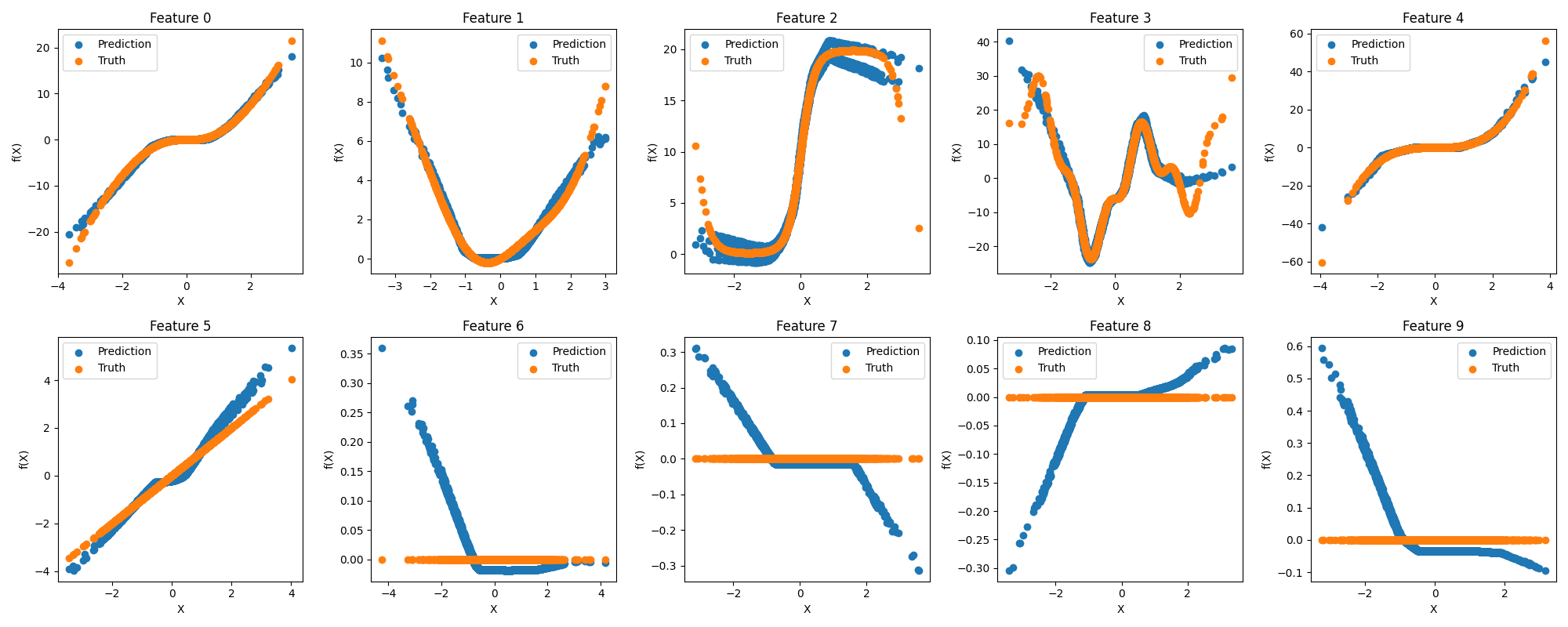}
    \includegraphics[width=.88\linewidth]{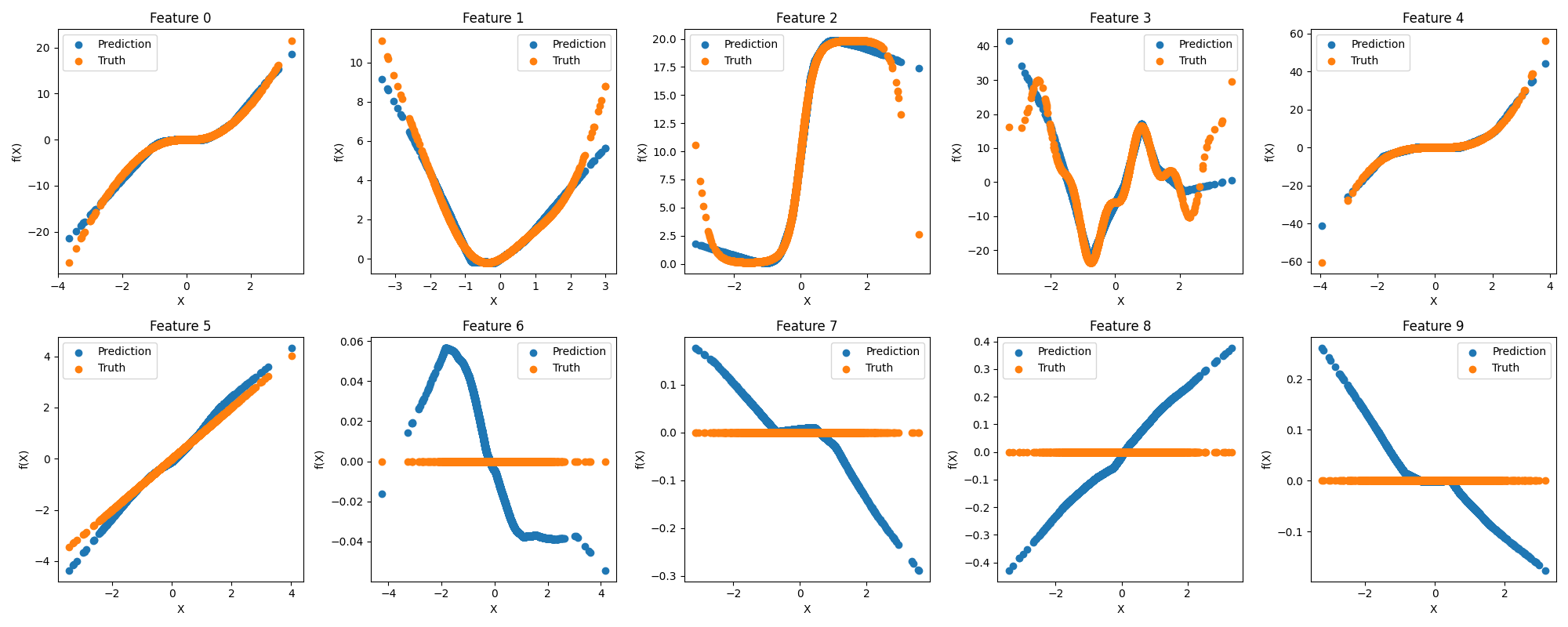}
    \includegraphics[width=.88\linewidth]{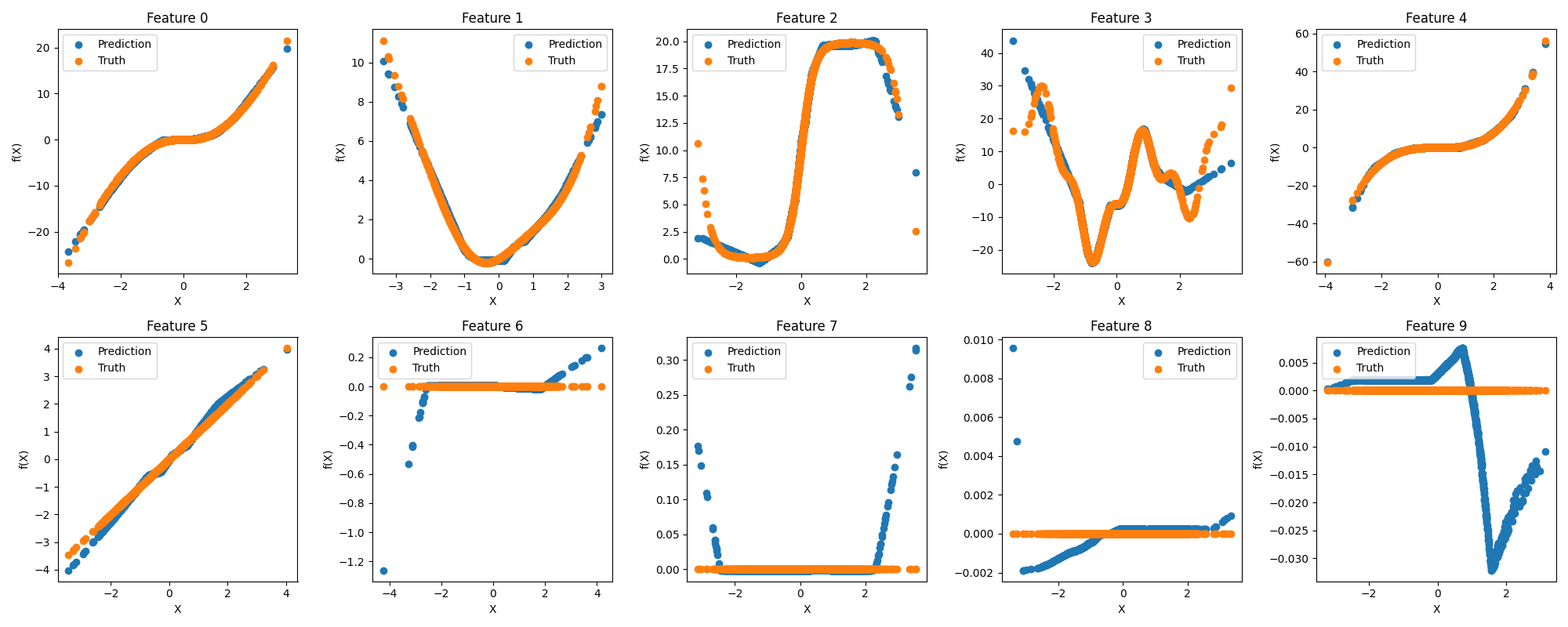}
    \caption{Individual effect learned by NAM on synthetic regression. Blue dots are predictions and orange dots are the truth. The first two rows are results optimized by ULR. The next two rows are features learned by Hi-ULR. The last two rows are the results of Hi-DLR.}
    \label{fig:nam-synthetic-features}
\end{figure}


\paragraph{Synthetic data}
The data $\bm X\in \R^{3000\times 10}$. Let's denote the $j$-th column of $\bm X$ as $\bm {X_j}$. $\y$ is generated by an additive model:
\begin{align*}
\y=\sum_{i=1}^{10}f_{i}(\bm {X_i})+\mathcal{N}(0,1) 
\end{align*}
where $f_j$ are zero functions for $j=7,8,9,10$. The rest features are generated in the following way:
\begin{align*}
    f_1(x) &= 2x^2\tanh{x}, \quad f_2(x) = \sin{x}\cos{x}+x^2
    , \quad f_3(x) = 20/(1+e^{-5\sin{x}})\\ 
    f_4(x) &= 20\sin^3{2x}-6\cos{x}+x^2, \quad
    f_5(x) = x^3, \quad f_6(x) = x
\end{align*}
For synthetic regression data, learning rates for ULR are selected from the list [5e-4, 7e-4, 1e-3, 3e-3, 5e-3, 7e-3, 1e-2]. All the models are trained with SGD. The total number of epochs is 100 and batch size is 256. $\Phi=2$ for Hi-ULR and Hi-DLR. Plots start from the 5th epoch.  

\paragraph{California housing}
This dataset collects the house values of various California districts in 1990. The regression task is to predict house prices with 20,640 examples and 8 housing features including location, layout, etc.

For California housing, learning rates for ULR are selected from a list [5e-6, 7e-6, 1e-5, 3e-5, 5e-5, 7e-5, 1e-4]. We use the Adam optimizer. The total number of epochs 200 is and batch size is 256. $\Phi=8$ for Hi-ULR and Hi-DLR. Plots start from the 50th epoch.

\clearpage

\section{Complexity analysis}\label{complexity analysis}
We follow the same analysis as in \cite{bu2024automatic} and it follows that Hi-DLR has the same peak memory cost as a base optimizer. For time complexity, we consider three operations: the forward pass $F$, the back-propagation $B$ and other costs $C$. Therefore, the base optimizer takes $F+B+C$ whereas Hi-DLR takes $(1+\frac{4K}{\Phi})F+B+C$. Here the additional computation is from extra forward passes. In a full-parameter training on a single GPU, $C$ is small and $B\approx 2F$, the relative training speed of Hi-DLR is $\geq\frac{1}{1+\frac{4K}{3\Phi}}$. For instance, when $K=2,\Phi=8$, Hi-DLR is roughly $\geq 75\%$ as fast as a base optimizer; when $K=2,\Phi=16$, Hi-DLR has a relative speed $\geq 86\%$. In practice, we observed a faster empirical speed than the theoretical speed, e.g. $\Phi=8$ actually has 81\% relative speed instead of the theoretical lower bound 75\% because of $C$. 
\begin{table}[!htb]
    \centering
    \begin{tabular}{c|c|c|c|c|c}
         $\Phi$ &$\infty$ (ULR) & 8 &4&2&1\\\hline
         Minutes& 8.5& 10.5 &12&17.5&26.5
    \end{tabular}
    \caption{The wall-clock time per epoch, training ViT-base on CIFAR100 with $K=2$. The experiment is performed on an A100 GPU.}
    \label{tab:my_label}
\end{table}

While training with PET methods, the $B\approx F$, the relative speed becomes $\geq \frac{1}{1+\frac{4K}{2\Phi}}$. When $K=2,\Phi=8$, Hi-DLR is $\geq 66.7\%$ as fast as a base optimizer.


\end{document}